\documentclass[11pt]{article}

\usepackage[a4paper,margin=1in]{geometry}

\usepackage[T1]{fontenc}
\usepackage[utf8]{inputenc}
\usepackage[english]{babel}
\usepackage{newtxtext}
\usepackage{newtxmath} % change to [amsthm] only if needed
\usepackage{microtype}
\usepackage{setspace}
\usepackage{mathtools}

\usepackage{graphicx}
\usepackage{booktabs}
\usepackage{multirow}
\usepackage{array}
\usepackage{caption}
\usepackage{subcaption}
\usepackage{float}

\usepackage[ruled,vlined,linesnumbered]{algorithm2e}

\usepackage[section]{placeins}
\usepackage{flafter}

\usepackage{enumitem}

\usepackage{authblk}

\usepackage[switch]{lineno}
\newif\ifdraftlinenumbers
\draftlinenumbersfalse
\usepackage{xcolor}
\definecolor{linkblue}{RGB}{0,82,155}
\usepackage[
    colorlinks=true,
    linkcolor=linkblue,
    citecolor=linkblue,
    urlcolor=linkblue
]{hyperref}
\usepackage[nameinlink,noabbrev]{cleveref}

\usepackage[authoryear]{natbib}

\usepackage{fancyhdr}
\usepackage{titlesec}
\titleformat{\section}{\large\bfseries}{\thesection.}{0.6em}{}
\titleformat{\subsection}{\normalsize\bfseries}{\thesubsection.}{0.6em}{}
\titleformat{\subsubsection}{\normalsize\itshape}{\thesubsubsection.}{0.5em}{}
\titlespacing*{\section}{0pt}{2.2ex plus .3ex minus .2ex}{1.0ex}
\titlespacing*{\subsection}{0pt}{1.8ex plus .2ex minus .2ex}{0.7ex}
\titlespacing*{\subsubsection}{0pt}{1.4ex plus .2ex minus .2ex}{0.5ex}

\usepackage{abstract}

\newcommand{\keywords}[1]{%
    \vspace{0.4em}
    \noindent\textbf{Keywords:} #1
}

\usepackage{xspace}
\usepackage{etoolbox}
\def\tsc#1{\csdef{#1}{\textsc{\lowercase{#1}}\xspace}}
\tsc{WGM}
\tsc{QE}

\title{\vspace{-1.2em}\bfseries\LARGE
Frenetic Cat-inspired Particle Optimization: a Markov state-switching hybrid swarm optimizer with application to cardiac digital twinning
}

\author[1]{Jorge S\'anchez\thanks{Corresponding author: \href{mailto:jorsana4@upv.edu.es}{jorsana4@upv.edu.es}}}
\author[1]{Guadalupe Garc\'ia-Isla}
\author[1]{Sandra Perez-Herrero}
\author[1]{Beatriz Trenor}
\author[1]{Javier Saiz}

\affil[1]{\small
Centro de Investigaci\'on e Innovaci\'on en Bioingenier\'ia (Ci2B), Universitat Polit\'ecnica de Val\`encia, Camino de Vera s/n, 46022 Valencia, Spain
}

\date{}

\begin{document}

\ifdraftlinenumbers
\linenumbers
\fi

\maketitle

\begin{abstract}
Designing optimizers that remain effective under tight evaluation budgets is critical in expensive black-box settings such as cardiac digital twinning. We propose \emph{Frenetic Cat-inspired Particle Optimization} (FCPO), a hybrid swarm method that couples particle swarm optimization-like dynamics with an explicit-state Markov switching controller to schedule exploration and refinement operators online. FCPO integrates (i) state-conditioned bounded motion, (ii) an elite-difference global jump operator to escape stagnation, (iii) eigen-space guided local refinement from elite covariance, and (iv) linear population size reduction to control late-stage computational cost. We benchmark FCPO on five representative functions from the Congress on Evolutionary Computation (CEC)~2022 suite (F1, F2, F3, F6 and F10) at dimensions $D\in\{10,20\}$ over 30 independent runs, comparing against PSO, CSO, CLPSO, SHADE, L-SHADE and CMA-ES. FCPO achieves the lowest mean runtime across the ten benchmark cases (average $0.183$\,s), about $2.3\times$ faster than CMA-ES and $2.6\times$ faster than L-SHADE in our Python implementation. On the multimodal composition function F10 at $D=20$, FCPO attains the best mean objective ($9.625\times 10^{2}\pm 1.275\times 10^{3}$) and remains faster than CMA-ES (0.602\,s vs.\ 1.126\,s mean runtime). On structured landscapes (F1--F3) and on the hybrid function (F6), CMA-ES remains the most accurate method, while FCPO substantially improves over classical swarms and maintains a favorable accuracy--runtime trade-off. Finally, in a ventricular activation digital twin calibration task, FCPO reaches the target electrocardiogram (ECG) fidelity (RMSE $<0.1$\,mV) within $\sim 40$ iterations and produces physiologically plausible activation maps with robust convergence across repeated initializations, supporting its use as a practical optimizer for expensive inverse problems.
\end{abstract}

% \begin{highlights}
% \item Markov state switching for online exploration--refinement scheduling
% \item Eigen-space refinement and population reduction improve speed--accuracy
% \item Benchmarked on CEC~2022 with a use case of ventricular activation digital-twin case study
% \end{highlights}


\vspace{1em}

%%%%%%%%%%%%%%%%%%%%%%%%%%
\section{Introduction}
Many scientific and engineering problems require optimizing black-box objectives that are non-convex, multimodal, and expensive to evaluate while remaining lightweight to execute~\citep{Jones1998}. In such settings, each function evaluation may involve a high-fidelity simulation, a complex numerical solve, or a costly experiment, so the central practical objective is not only solution quality but also minimizing the number of evaluations and wall-clock time. Population-based metaheuristics are attractive in this regime because they require limited problem structure and can handle irregular landscapes; however, they often struggle with stagnation, sensitivity to hyperparameters, or high evaluation budgets. This motivates hybrid designs that combine fast swarm-style search with occasional diversity injection and principled local refinement, while remaining lightweight to execute.

Particle Swarm Optimization (PSO) is a canonical example of a lightweight population method that can make rapid early progress on many problems~\citep{Kennedy1995}. Yet PSO can stagnate prematurely on complex multimodal and ill-conditioned landscapes, motivating variants that enhance diversity and learning mechanisms~\citep{Liang2006}. Differential Evolution (DE) provides strong global search behavior through differential mutation and recombination~\cite{Storn1997}, and modern DE variants incorporate adaptation and population size control to improve robustness and efficiency~\citep{Zhang2009,Tanabe2013,Tanabe2014}. Evolution strategies such as Covariance Matrix Adaptation Evolution Strategy (CMA-ES) further improve performance on ill-conditioned problems by adapting a covariance model of the search distribution~\citep{HansenOstermeier2001,Hansen2016}, but can incur higher per-iteration overhead and may be less attractive when the evaluation budget is tight and implementation simplicity is a priority. Since no optimizer can dominate across all problem classes~\citep{Wolpert1997}, progress in this area is best achieved by clearly specifying the intended operating regime and rigorously evaluating on diverse benchmarks.

A recurring theme in successful hybrid metaheuristics is episodic exploration: maintaining efficient exploitation dynamics most of the time, while enabling intermittent high-energy transitions that can escape stagnation. Metaphor-inspired algorithms often describe such switching behavior, but convincing contributions require operator-level clarity and empirical evidence rather than narrative novelty~\citep{Sorensen2015}. Cat Swarm Optimization (CSO)~\citep{Chu2006} is an early example of a mode-switching swarm inspired by feline behavior; for a physics-based discussion of cat motion as a dynamical system, see also~\citep{Biasi2024}.

In this work we propose \emph{Frenetic Cat-inspired Particle Optimization (FCPO)}, a hybrid swarm framework that formalizes \emph{stochastic, explicit-state switching} between exploitation and exploration operators using a Markov switching controller. FCPO couples three mechanisms intended for expensive, multimodal objectives: (i) a PSO-like baseline update for efficient progress~\citep{Kennedy1995}, (ii) an exploration ``zoomies'' operator based on elite-difference jumps inspired by DE mutation principles~\citep{Storn1997}, and (iii) covariance-informed Eigen-space guidance, updated from elite solutions, to provide anisotropic refinement directions motivated by covariance adaptation ideas~\cite{HansenOstermeier2001,Hansen2016}. To reduce late-stage evaluation cost, FCPO further integrates Linear Population Size Reduction (LPSR)~\citep{Tanabe2014}. The Markov transition matrix is adapted online to reinforce behaviors associated with the current best particle and to bias transitions toward exploration under stagnation, yielding a heterogeneous population with state-dependent dynamics.

Although FCPO is proposed as a general-purpose optimizer, we use cardiac digital twinning as a motivating use case of the target regime: high-dimensional, ill-posed inverse calibration under strict evaluation budgets. Recent ECG-driven personalization workflows combine fast electrophysiology surrogates (e.g., eikonal-based activation) with efficient forward operators to enable optimization-in-the-loop model fitting, yet the resulting objective landscapes remain multimodal and prone to stagnation when only a limited number of forward evaluations is feasible~\citep{Grandits2021GEASI,Grandits2020InverseEikonal,Gillette2021,Camps2021,Camps2024PurkinjeTwin,Grandits2025}. We therefore apply FCPO to the creation of cardiac digital twins as a representative high-cost calibration scenario.

We evaluate FCPO on representative functions from the CEC 2022 single-objective bound-constrained benchmark suite~\citep{Kumar2021}, using multiple dimensions and repeated trials, and report both solution quality and computational cost. Because different metaheuristics can perform different numbers of objective evaluations per iteration, we additionally report the number of function evaluations and runtimes to contextualize comparisons. Finally, to illustrate the target application regime of expensive inverse problems, we demonstrate FCPO on a cardiac activation calibration task from surface ECGs; this example serves as a motivating case study rather than the primary contribution of the paper.

\subsection{Related work and positioning}
\label{sec:related}

FCPO sits at the intersection of swarm intelligence, adaptive hybrid metaheuristics, and state-switching control. To position the method clearly, we summarize the closest lines of work and the specific gaps they leave in the low-evaluation and high-cost regime targeted here.

\paragraph{PSO variants and heterogeneous swarms.}
PSO is a standard low-overhead baseline for black-box search~\citep{Kennedy1995}. Many PSO variants improve robustness on multimodal landscapes via modified learning strategies (e.g., comprehensive learning), dynamic parameters (e.g., inertia schedules), or multi-swarm/diversity mechanisms~\citep{Liang2006}. These methods typically retain a single continuous update operator (a PSO-style velocity rule) and modulate it through parameter adaptation or topology changes. In contrast, FCPO assigns each particle an explicit discrete behavioral state that selects among multiple qualitatively different operators (PSO step, elite-difference jump, eigen-aligned refinement, damped restoration), thereby producing a heterogeneous swarm without manually partitioning the population.

\paragraph{Adaptive operator selection and hybrid metaheuristics.}
A broad class of hybrids dynamically chooses among multiple search operators (e.g., mutation/crossover families in DE or local-search kernels), often through reward-based adaptive operator selection or bandit-style credit assignment~\citep{Fialho2010,Maturana2012}. FCPO is related in spirit—operator usage adapts online—but differs in mechanism and granularity: operator choice is implemented as a Markov switching controller over a small state space, and the transition matrix is reinforced toward the state of the current best particle (and biased toward exploration under stagnation). This makes the adaptation explicit, stateful, and particle-wise.

\paragraph{Markov-switching metaheuristics and state-transition control.}
State-transition ideas have appeared in optimization, where search phases (exploration vs.\ exploitation) are represented as discrete modes. However, many approaches either (i) prescribe deterministic phase schedules, (ii) use global mode switches, or (iii) employ static transition rules. FCPO uses an explicit-state, particle-wise Markov chain whose transition matrix is adapted online using simple reinforcement signals (best-state reinforcement and stagnation-triggered exploration bias), enabling each particle to follow its own stochastic behavioral trajectory. This controller is computationally negligible compared to expensive objectives and is designed for limited evaluation budgets.

\paragraph{Differential Evolution variants and modern baselines.}
DE remains a competitive family for continuous optimization; modern adaptive variants such as JADE, SHADE and L-SHADE achieve strong performance via parameter adaptation and population-size control~\citep{Zhang2009,Tanabe2013,Tanabe2014}. FCPO borrows one DE-like ingredient—an elite-difference jump operator (Zoomies)—but integrates it in a swarm framework where (1) DE-style jumps are episodic and governed by Markov switching, and (2) late-stage computation is reduced by LPSR while local refinement is guided by an eigen-system learned from elites (conceptually related to covariance-informed search used in CMA-ES)~\citep{Hansen2016}. The resulting objective is not to outperform DE/CMA-ES in asymptotic best value, but to obtain a favorable accuracy--runtime trade-off when evaluations are costly and budgets are tight.

\paragraph{Summary of the gap.}
Existing PSO variants often improve robustness but still stagnate on hybrid/composition landscapes under strict budgets, while modern DE/CMA-ES methods can be more evaluation-intensive or incur higher per-iteration overhead. FCPO targets this gap by combining (i) a low-cost PSO backbone, (ii) particle-wise state switching that injects exploration only when needed, (iii) LPSR to reduce late-stage evaluation cost, and (iv) structured eigen-aligned refinement to stabilize fine-tuning.

\section{Methods}\label{sec2}

\begin{figure}[!t]
    \centering
    \includegraphics[width=0.5\textwidth]{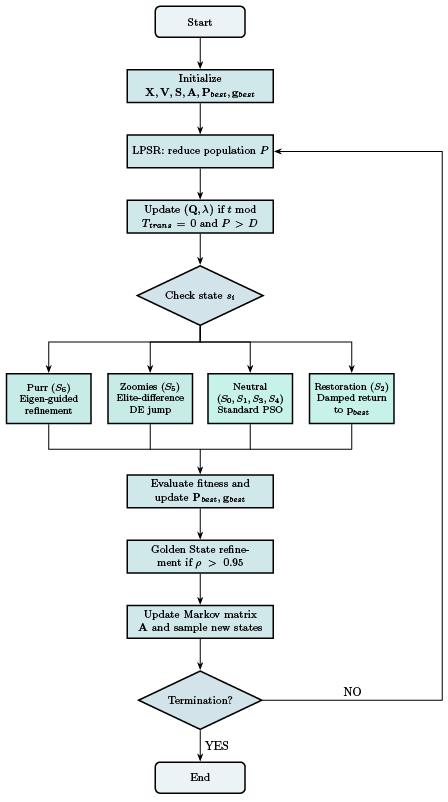}
    \caption{The algorithm iteratively combines population reduction, state-dependent particle updates, fitness evaluation, late-stage refinement, and Markov-state transitions until termination. In the figure, \(\mathbf{X}\) and \(\mathbf{V}\) denote particle positions and velocities, \(\mathbf{S}\) the vector of particle states, \(\mathbf{A}\) the Markov transition matrix, \(\mathbf{P}_{best}\) the personal-best positions, \(\mathbf{g}_{best}\) the global-best solution, \(P\) the current population size, \(D\) the number of decision variables, \((\mathbf{Q},\boldsymbol{\lambda})\) the elite covariance eigensystem, \(t\) the current iteration, \(T_{\text{trans}}\) the transition period, and \(\rho=t/T_{\max}\) the normalized iteration progress.}
    \label{fig:fcpo_diagram}
\end{figure}

The core of \emph{Frenetic Cat-inspired Particle Optimization (FCPO)} is summarized in the flowchart in Figure~\ref{fig:fcpo_diagram} and formalized in Algorithm~\ref{alg:fcpo_algorithm}. Each particle represents an individual search agent whose motion is governed by a baseline PSO update and intermittently modified by state-conditioned operators. The initial population is generated by Latin Hypercube Sampling (LHS)~\citep{McKay1979} using a maximin space-filling criterion~\citep{MorrisMitchell1995} over multiple candidate designs to improve coverage of the search domain. %Unless specified otherwise, the initial population size is set to $P_0=10D$.
Unless specified otherwise, the initial population size is set to $P_0=10D$, i.e. 10 particles per decision variable, where $D$ denotes the problem dimension.

To reduce evaluation cost in later stages, FCPO applies LPSR~\citep{Tanabe2014}, progressively shrinking the swarm from $P_{\text{init}}$ to $P_{\min}$. In the baseline (\emph{neutral}) behavioral states, particles follow the standard PSO mechanics~\citep{Kennedy1995,Shi1998,Clerc2002}: velocities are updated by combining inertia with a cognitive pull toward each particle's personal best ($\mathbf{p}_{best}$) and a social pull toward the global best ($\mathbf{g}_{best}$). To incorporate landscape geometry, FCPO periodically estimates a covariance matrix from an elite subset of personal-best solutions and uses its eigensystem to guide anisotropic motion and perturbations, conceptually akin to covariance-informed search in CMA-ES~\citep{HansenOstermeier2001,Hansen2016}. 
%In implementation, velocities are clipped component-wise to $\mathbf{v}_{\max}=0.2(\mathbf{ub}-\mathbf{lb})$ to stabilize early exploration. The inertia weight follows a cosine schedule $w(\rho)=0.4+0.5\cos(\pi\rho)$ with $w(\rho)\ge 0.1$; during the final 2\% of iterations ($\rho>0.98$), a terminal lockdown sets $w=0$ and reduces the velocity bound to $10^{-6}\mathbf{v}_{\max}$ to suppress late oscillations~\cite{Shi1998}.
Let \(\rho=t/T_{\max}\) denote the normalized iteration progress, where \(t\) is the current iteration and \(T_{\max}\) is the maximum number of iterations, and let \(\mathbf{lb},\mathbf{ub}\in\mathbb{R}^D\) denote the lower and upper bound vectors of the search space. In our implementation, velocities are clipped component-wise to \(\mathbf{v}_{\max}=0.2(\mathbf{ub}-\mathbf{lb})\) to stabilize early exploration. The inertia weight follows the cosine schedule \(w(\rho)=0.4+0.5\cos(\pi\rho)\), with \(w(\rho)\ge 0.1\). During the final \(2\%\) of iterations (\(\rho>0.98\)), a terminal lockdown sets \(w=0\) and reduces the velocity bound to \(10^{-6}\mathbf{v}_{\max}\) to suppress late oscillations~\citep{Shi1998}.

A distinctive feature of FCPO is its explicit-state Markov switching controller: each particle carries a discrete state that selects an operator at each iteration. For most of the runtime—specifically in the neutral states $S_0,S_1,S_3,$ and $S_4$—particles remain in a ``stalking'' mode and apply only the baseline PSO update. In contrast, special states trigger ethology-inspired overrides. The \emph{zoomies} state $S_5$ executes a DE-style elite-difference jump~\citep{Storn1997}, using elites selected from the best-ranked personal bests (in the spirit of ``$p$-best'' guidance used in adaptive DE variants such as JADE, SHADE and L-SHADE~\citep{Zhang2009,Tanabe2013,Tanabe2014}). The \emph{purr} state $S_6$ performs Eigen-aligned local refinement around $\mathbf{p}_{best}$ using the elite covariance eigensystem (CMA-ES-like directional structure~\citep{Hansen2016}), while the \emph{restoration} state $S_2$ damps motion and pulls the particle back toward $\mathbf{p}_{best}$ to mitigate overshoot.

% Concurrently, the algorithm updates the elite eigensystem every $T_{\text{trans}}$ iterations (when $P>D$) and uses it to guide a multi-scale ``Golden State'' local search near the end of the run. By combining LPSR~\cite{Tanabe2014}, PSO baseline dynamics~\cite{Kennedy1995}, covariance-informed directional guidance~\cite{Hansen2016}, and episodic DE-type jumps~\cite{Storn1997,Zhang2009}, FCPO aims to maintain a favorable accuracy--runtime trade-off on expensive and multimodal objectives.

Concurrently, every $T_{\text{trans}}$ iterations, FCPO recomputes the eigensystem of the covariance matrix estimated from an elite subset of personal-best solutions, provided that the current population size $P$ remains larger than the search-space dimension $D$. Here, $P$ denotes the current number of particles, while $D$ denotes the number of decision variables. This condition ensures that the covariance-based directional information is estimated only when the swarm is still large enough to provide a meaningful representation of the local search geometry. Near the end of the run, this eigensystem is further used to guide the multi-scale ``Golden State'' local search. By combining LPSR~\cite{Tanabe2014}, PSO baseline dynamics~\cite{Kennedy1995}, covariance-informed directional guidance~\cite{Hansen2016}, and episodic DE-type jumps~\cite{Storn1997,Zhang2009}, FCPO aims to maintain a favorable accuracy--runtime trade-off on expensive and multimodal objectives.

\begin{algorithm}[!htpb]
\small
\DontPrintSemicolon
\caption{Frenetic Cat-inspired Particle Optimization (FCPO)}
\label{alg:fcpo_algorithm}

\KwIn{Objective function $f(\mathbf{x})$, dimension $D$, bounds $[\mathbf{lb},\mathbf{ub}]$, maximum iterations $T_{\max}$.}
\KwOut{Best solution $\mathbf{g}_{best}$ and its objective value $J_g$.}

Initialize particle positions $\mathbf{X}$ inside bounds\;
Initialize velocities $\mathbf{v}$\;
Set each particle's personal best $\mathbf{P}_{best}\leftarrow \mathbf{X}$\;
Evaluate objective values $\mathbf{J}_{best}\leftarrow f(\mathbf{X})$\;
Set global best $\mathbf{g}_{best}$ and $J_g$\;
Initialize particle states $\mathbf{s}$ and transition matrix $\mathbf{A}$\;

\For{$t \leftarrow 1$ \KwTo $T_{\max}$}{

    Compute progress ratio $\rho = t/T_{\max}$\;

    Reduce population gradually if needed\;

    Update particle velocities using:
    \begin{itemize}
        \item inertia,
        \item attraction to personal best,
        \item attraction to global best
    \end{itemize}

    Limit velocities and update positions\;
    Keep positions inside bounds\;

    \If{$t$ is a transition step}{
        Update search directions from current best particles\;
    }

    Apply state-dependent search operators:
    \begin{itemize}
        \item State 5: exploratory differential move using elite particles
        \item State 6: local Gaussian search around personal best
        \item State 2: partial return toward personal best
    \end{itemize}

    Keep positions inside bounds\;
    Evaluate new objective values $\mathbf{J} = f(\mathbf{X})$\;

    Update personal bests where improvement occurs\;
    Update global best if a better solution is found\;

    \If{search is near the end}{
        Apply Golden State refinement to $\mathbf{g}_{best}$\;
    }

    \If{$t$ is a transition step}{
        Update transition matrix $\mathbf{A}$ based on progress\;
        Sample new particle states $\mathbf{s}$ from $\mathbf{A}$\;
    }
}

\Return{$\mathbf{g}_{best}, J_g$}\;
\end{algorithm}

\section{Markov Switching Transition States}

The behavioral dynamics of the $i$-th particle at iteration $t$ are governed by an explicit-state Markov switching controller, i.e., a discrete-time Markov chain over a finite set of behavioral modes. Each particle stores its current state and the state directly selects which update operator is applied at that iteration.

\subsection{State Space}
We consider $N=7$ discrete states
\begin{equation}
    \mathcal{S} = \{ S_0, S_1, S_2, S_3, S_4, S_5, S_6 \},
\end{equation}
grouped into four operational categories:
\begin{itemize}
    \item \textbf{Neutral Maintenance} ($\mathcal{S}_{neutral} = \{ S_0, S_1, S_3, S_4 \}$): latent neutral substates that all apply the same PSO-based movement operator, but remain separated in the Markov controller to allow different transition patterns toward exploration, restoration, and refinement.
    \item \textbf{Restoration} ($S_2$): damped pull-back toward the personal best (stability).
    \item \textbf{Exploration} ($S_5$): DE-inspired elite-difference jump (escape from stagnation).
    \item \textbf{Stabilization} ($S_6$): Eigen-aligned local refinement around the personal best (ridge-following).
\end{itemize}

Although \(S_0\), \(S_1\), \(S_3\), and \(S_4\) all apply the same PSO-based update, they are treated as separate neutral substates in the Markov controller. This does not change the instantaneous particle motion, but enriches the transition structure by allowing different pathways from neutral behavior toward restoration, exploration, or local refinement. In this sense, these states are dynamically identical at the update level, but distinct in how they participate in the state-transition process.

\subsection{Markov Transition Mechanism}
Let $s_i^{(t)} \in \mathcal{S}$ denote the explicit state of particle $i$ at iteration $t$. Transitions are governed by a row-stochastic matrix $\mathbf{A}\in\mathbb{R}^{7\times 7}$:
\begin{equation}
    a_{jk} = P\!\left(s_i^{(t+1)} = S_k \mid s_i^{(t)} = S_j\right), 
    \qquad \sum_{k=0}^{6} a_{jk} = 1 \;\; \forall j.
\end{equation}
The next state is sampled as
\begin{equation}
    s_i^{(t+1)} \sim \mathrm{Categorical}\!\left(\mathbf{A}_{s_i^{(t)},:}\right).
\end{equation}

In the implementation, $\mathbf{A}$ is initialized uniformly and then adapted every $T_{\text{trans}}$ iterations. Let $g$ denote the index of the current best particle and $s^\ast = s_g^{(t)}$ its state. A column-reinforcement update is applied to all rows with $\eta=0.2$:
\begin{equation}
A_{r,s^\ast} \leftarrow (1-\eta)A_{r,s^\ast}+\eta,\qquad \forall r\in\{0,\dots,6\},
\end{equation}
and under stagnation ($no\_imp>10$) the exploration state is encouraged by
\begin{equation}
A_{r,5} \leftarrow A_{r,5}+0.4,\qquad \forall r.
\end{equation}
Each row is then renormalized to sum to one before sampling the next states.

\subsection{State-Conditioned Dynamics}
Let $\mathbf{x}_i^{(t)}\in\mathbb{R}^D$ and $\mathbf{v}_i^{(t)}\in\mathbb{R}^D$ denote the position and velocity of particle $i$ at iteration $t$. Let $\mathbf{p}_{best,i}$ be its personal best and $\mathbf{g}_{best}$ the global best. The progress ratio is $\rho=t/T_{\max}$. For all updates, positions are clipped to the bound constraints.

\subsubsection{Neutral Maintenance ($s_i^{(t)} \in \mathcal{S}_{neutral}$)}
Particles in neutral states follow the standard PSO velocity update:
\begin{align}
    \mathbf{v}_i^{(t+1)} &= w(\rho)\,\mathbf{v}_i^{(t)} 
    + c_1 r_1 \left(\mathbf{p}_{best,i} - \mathbf{x}_i^{(t)}\right)
    + c_2 r_2 \left(\mathbf{g}_{best} - \mathbf{x}_i^{(t)}\right), \\
    \mathbf{x}_i^{(t+1)} &= \mathbf{x}_i^{(t)} + \mathbf{v}_i^{(t+1)}.
\end{align}
Here $\mathbf{r}_1,\mathbf{r}_2\sim\mathcal{U}(0,1)^D$ are independent random vectors sampled component-wise. In the implementation, $w(\rho)=0.4+0.5\cos(\pi\rho)$ with $w(\rho)\ge 0.1$, and velocities are clipped component-wise to $\pm v_{\max}$, with a terminal lockdown for $\rho>0.98$ (Algorithm~\ref{alg:fcpo_algorithm}).

\subsubsection{Exploration State: ``Zoomies'' ($s_i^{(t)} = S_5$)}
During the global exploration phase (implemented for $\rho<0.9$), this state applies a DE-inspired elite-difference jump using personal bests. Let $\mathcal{E}$ be the elite set defined as the top $K=\max(2,\lfloor 0.4P\rfloor)$ particles ranked by personal-best objective value. Choose two elites $a,b\in\mathcal{E}$ and reorder them such that $a$ is better than $b$, then apply
\begin{align}
    \mathbf{x}_i^{(t+1)} &= \mathbf{p}_{best,a} + F\left(\mathbf{p}_{best,a} - \mathbf{p}_{best,b}\right), \\
    \mathbf{v}_i^{(t+1)} &= \mathbf{0},
\end{align}
where $F\sim\mathcal{N}(0.5,0.3)$ in the implementation. This jump helps break stagnation by moving along a direction defined by elite differences.

\subsubsection{Stabilization State: ``Purr'' ($s_i^{(t)} = S_6$)}
This state performs Eigen-aligned local refinement around the personal best. Let $\mathbf{C}$ be the covariance of the elite set with eigendecomposition $\mathbf{C} = \mathbf{Q}\mathbf{\Lambda}\mathbf{Q}^T$ and eigenvalues $\boldsymbol{\lambda}=\mathrm{diag}(\mathbf{\Lambda})$. The implementation samples $\mathbf{\Xi}\sim\mathcal{N}(\mathbf{0},\mathbf{I}_D)$ and applies a normalized eigenvalue scaling
\begin{equation}
\tilde{\mathbf{s}}=\frac{\sqrt{\boldsymbol{\lambda}+10^{-10}}}{\max_k\sqrt{\lambda_k+10^{-10}}},
\end{equation}
followed by rotation back to world coordinates and scaling by the parameter range:
\begin{align}
    \mathbf{x}_i^{(t+1)} &= \mathbf{p}_{best,i} + \alpha(\rho)\,
    \left[\mathbf{Q}\left(\tilde{\mathbf{s}}\odot \mathbf{\Xi}\right)\right]\odot(\mathbf{ub}-\mathbf{lb}), \\
    \mathbf{v}_i^{(t+1)} &= \mathbf{0},
\end{align}
with $\alpha(\rho)=0.02(1-\rho)^2$.

\subsubsection{Restoration State: ``Oscillation'' ($s_i^{(t)} = S_2$)}
This state implements a simple damped pull-back toward the personal best:
\begin{align}
    \mathbf{v}_i^{(t+1)} &= 0.5\,\mathbf{v}_i^{(t)}, \\
    \mathbf{x}_i^{(t+1)} &= \mathbf{x}_i^{(t)} + 0.5\left(\mathbf{p}_{best,i}-\mathbf{x}_i^{(t)}\right).
\end{align}
This operator reduces overly aggressive trajectories and promotes stability when particles drift away from previously good regions.

\section{Benchmarking}\label{sec:benchmarking}

To evaluate FCPO, we benchmark it against widely used optimizers spanning swarm intelligence, differential evolution, and evolution strategies: PSO~\citep{Kennedy1995}, CSO~\cite{Chu2006}, Comprehensive Learning PSO (CLPSO)~\citep{Liang2006}, SHADE~\cite{Tanabe2013}, L-SHADE~\citep{Tanabe2014}, and Covariance Matrix Adaptation Evolution Strategy (CMA-ES)~\citep{Hansen2016}. The evaluation protocol combines per-case nonparametric significance testing with aggregate rank-based analysis in order to assess both local and cross-problem performance differences. The experimental protocol and reporting choices were guided by recent recommendations for benchmarking bio-inspired optimization algorithms and for statistically comparing stochastic optimizers~\citep{Latorre2021,Carrasco2020,Derrac2011}. This set includes classical swarm baselines (PSO/CSO), improved swarm learning dynamics (CLPSO), state-of-the-art adaptive DE variants (SHADE/L-SHADE), and a strong covariance-based strategy (CMA-ES).

\subsection{Benchmark suite and problem selection}
We used the CEC~2022 single-objective bound-constrained benchmark suite~\citep{Kumar2021} and selected five functions to cover distinct landscape classes while keeping the analysis interpretable: the unimodal Zakharov function (CEC22-F1), the shifted and rotated Rosenbrock function (CEC22-F2), the expanded Schaffer’s F7 function (CEC22-F3), the hybrid function (CEC22-F6), and a multimodal composition function (CEC22-F10). These were chosen to expose complementary difficulty profiles (unimodal, curved-valley under rotation, rugged near-optimum behavior, hybrid structure, and composition multimodality). Experiments were conducted at dimensions $D\in\{10,20\}$ using the CEC~2022 function definitions and bounds~\citep{Kumar2021}. Each algorithm was run for 30 independent trials with different random seeds.

\subsection{Stopping rule, function-evaluation accounting, and reported metrics}
All algorithms were evaluated under a matched maximum function-evaluation budget. Because different metaheuristics can realize different internal trajectories and convergence patterns even under the same evaluation limit, we recorded for every run: (i) the final best objective value at termination, (ii) the best-so-far trajectory, (iii) the realized number of function evaluations (NFE), and (iv) the wall-clock runtime measured with identical timing instrumentation. Convergence is therefore reported versus realized NFE, and runtime is reported as the total wall-clock time required by each optimizer under the shared benchmark protocol and computational environment.

For each function--dimension case, every algorithm was run 30 times with independent random seeds. We report the mean and standard deviation of the final best objective values in the statistical analysis. Full descriptive statistics for all methods are provided in the Supplementary Tables~\ref{tab:benchmark_stats_d10}--\ref{tab:benchmark_stats_d20}.

\subsection{Algorithm configurations}
For population-based methods, PSO, CSO, CLPSO, and FCPO were initialized with $N=30$ particles. In FCPO, linear population size reduction (LPSR) was further applied from $P_{\mathrm{init}}=30$ down to $P_{\min}=4$. For the CEC~2022 experiments, we overrode the default FCPO setting $P_0=10D$ and used a fixed initial population size $N=30$ for all swarm-based methods in order to standardize the computational budget and avoid conflating algorithmic behavior with different default swarm sizes. We acknowledge that this departs from the default FCPO recommendation, and therefore the reported results should be interpreted as comparisons under a standardized-budget protocol rather than fully default-tuned settings for every method. Accordingly, this design choice is discussed as a fairness--fidelity trade-off. CMA-ES used the standard recommended settings in our implementation, whereas SHADE and L-SHADE were run with their standard adaptive parameter and population-control mechanisms.

\subsection{Implementation and computational environment}
All optimizers were implemented in Python. Benchmark functions were evaluated using the Opfunu library~\cite{Opfunu}. Experiments were executed on a workstation equipped with a 13th Gen Intel(R) Core(TM) i7-13700KF and 64\,GB RAM under Linux, using a single CPU process. The FCPO implementation is available at: \url{https://github.com/jorge221/fcpo}.

\subsection{Statistical analysis}
The statistical analysis follows current recommendations for comparing stochastic optimization algorithms~\cite{Derrac2011,Carrasco2020,Latorre2021}. Following these recommendations, we use 30 independent runs per function--dimension case, report both central tendency and dispersion, avoid relying solely on mean and standard deviation summaries, and combine nonparametric within-case tests with rank-based aggregate analyses across cases to reduce sensitivity to non-Gaussian and heteroscedastic performance distributions. For each function--dimension case, we first tested for overall differences among algorithms using a Kruskal--Wallis test on the final best objective values obtained over 30 independent runs. When the omnibus test rejected the null hypothesis of equal performance, post-hoc pairwise comparisons against FCPO were performed using Dunn's test with Holm correction for multiple testing.

To complement significance testing, we also report Cliff's delta as a nonparametric effect-size measure for the pairwise comparisons against FCPO. For minimization problems, a negative Cliff's delta indicates that FCPO tends to achieve lower objective values than the comparator, whereas a positive value indicates the opposite. Effect sizes are interpreted using the standard negligible, small, medium, and large categories.

At the aggregate level, overall performance across the 10 benchmark cases was summarized using per-case ranks computed from the median final objective value of each algorithm, with lower ranks indicating better performance. Average ranks were then computed across cases, and a Friedman test was applied to the case-level ranks of algorithms with complete coverage across all benchmark instances. This aggregate analysis complements the per-case hypothesis tests by summarizing relative performance across problems rather than within a single benchmark instance.

\section{Ventricular Activation Digital-Twin Calibration}
In human cardiac electrophysiology, ventricular activation is a complex, rapid process initiated at discrete sites known as Purkinje-Myocyte Junctions (PMJs). To reconstruct this activation pattern, we model the PMJs as particles within the FCPO algorithmic framework. The inverse problem is formulated as finding the optimal set of parameters $\mathbf{x}$ that minimizes the dissimilarity between the clinical target ECG ($Y_{target}$) and the simulated ECG ($Y_{sim}$). All data used in the case study were acquired in accordance with the Declaration of Helsinki, with approval from the relevant institutional ethics committee, and with informed consent obtained from all participants or waived as appropriate by the approving institution.

\subsection{Forward Model: Anisotropic Eikonal activation and lead-field ECG}
Simulations of ventricular activation were performed in openCARP~\cite{openCARP-paper,openCARP-sw} using an anisotropic eikonal model~\cite{Gillette2021,Grandits2020InverseEikonal,BarriosEspinosa2025}, in which the wavefront arrival time $t_a(\mathbf{x})$ satisfies
\begin{equation}
\nabla t_a(\mathbf{x})^{T}\,\mathbf{V}(\mathbf{x})\,\nabla t_a(\mathbf{x}) = 1,
\qquad \mathbf{x}\in\Omega_{biv},
\end{equation}
with Dirichlet activation times prescribed at initiation sites (e.g., Purkinje--myocyte junction locations).
Following the formulation in \cite{Gillette2021}, the tensor field $\mathbf{V}$ encodes the squared orthotropic conduction velocities,
\begin{equation}
\mathbf{V}(\mathbf{x}) =
c_f(\mathbf{x})^2\,\mathbf{f}(\mathbf{x})\mathbf{f}(\mathbf{x})^{T} +
c_s(\mathbf{x})^2\,\mathbf{s}(\mathbf{x})\mathbf{s}(\mathbf{x})^{T} +
c_n(\mathbf{x})^2\,\mathbf{n}(\mathbf{x})\mathbf{n}(\mathbf{x})^{T},
\end{equation}
where $\mathbf{f},\mathbf{s},\mathbf{n}$ denote the fiber, sheet, and sheet-normal directions.

To compute ECGs efficiently, we use a lead-field approach. For each electrode location $e$ (with reference electrode at $r$), the lead field $Z(\mathbf{x})$ is obtained by solving
\begin{equation}
\nabla\cdot\big((\sigma_i+\sigma_e)\nabla Z(\mathbf{x})\big)=
I_e\,\delta(\mathbf{x}-e) - I_r\,\delta(\mathbf{x}-r),
\end{equation}
with unit currents $|I_e|=|I_r|=1$ and intracellular/extracellular conductivity tensors $\sigma_i,\sigma_e$.
The extracellular potential at the electrode is then computed as
\begin{equation}
\phi_e(e,t)=\int_{\Omega_{biv}} \nabla Z(\mathbf{x})\cdot \sigma_i(\mathbf{x})\,\nabla V_m(\mathbf{x},t)\,d\Omega,
\end{equation}
where $V_m(\mathbf{x},t)$ is the transmembrane potential generated by the chosen source model.
As the right-hand side of the lead-field equation is constant, $Z$ can be computed once during initialization and reused, making ECG evaluation per iteration inexpensive~\cite{Potse2018,Neic2017}.

\subsection{Parameterization}
The optimization space is defined by the number of PMJs ($N_{PMJ}$) junction sites. Each PMJ $i$ is characterized by a parameter vector $\mathbf{p}_i = \{ab_i, rt_i, tm_i, t_{onset, i}\}$, representing its spatial coordinates using a universal coordinate system and its activation onset time.

\subsection{Objective function: \(L_2\) loss with global alignment, scaling, and offset}
Similarity between the measured ECG \(Y_{\mathrm{target}}\) and simulated ECG \(Y_{\mathrm{sim}}\) is quantified using an \(L_2\) loss. Because a direct \(L_2\) comparison is sensitive to temporal shifts and amplitude differences, the signals are first aligned in time and then jointly scaled following \cite{Gillette2021}. The same time shift, scale, and offset are applied across all 12 leads to preserve inter-lead relationships.

Let \(E\) denote the set of 12 leads, and let \(Y_{\mathrm{target},l}(t)\) and \(Y_{\mathrm{sim},l}(t)\) denote the measured and simulated signals for lead \(l\in E\). For each lead \(l\), the dominant peak times are defined as
\begin{equation}
t_{\mathrm{target},l} = \arg\max_{t\in[0,T_{\mathrm{target}}]} Y_{\mathrm{target},l}(t)^2,
\qquad
t_{\mathrm{sim},l} = \arg\max_{t\in[0,T_{\mathrm{sim}}]} Y_{\mathrm{sim},l}(t)^2.
\end{equation}
A candidate time shift \(\delta t_l\) is then defined by
\begin{equation}
\delta t_l = t_{\mathrm{target},l} - t_{\mathrm{sim},l}.
\end{equation}
Given \(\delta t_l\), the optimal scale \(s_l\) and offset \(r_l\) for lead \(l\) are defined as the minimizers of
\begin{equation}
\min_{s,r\in\mathbb{R}}
\int_{0}^{T_{\mathrm{sim}}}\left|Y_{\mathrm{target},l}(t+\delta t_l) - \big(s\,Y_{\mathrm{sim},l}(t)+r\big)\right|^2\,dt.
\label{eq:l2_scale_offset}
\end{equation}

Among the 12 candidate tuples \((\delta t_l,s_l,r_l)\), the optimal tuple \((\delta t_{\mathrm{opt}},s_{\mathrm{opt}},r_{\mathrm{opt}})\) is selected according to
\begin{equation}
l_{\mathrm{opt}}=\arg\min_{l\in E}
\left(
\sum_{k\in E}\int_{0}^{T_{\mathrm{sim}}}
\left|Y_{\mathrm{target},k}(t+\delta t_l) - \big(s_l\,Y_{\mathrm{sim},k}(t)+r_l\big)\right|^2\,dt
\right),
\end{equation}
and the final loss is
\begin{equation}
\mathcal{L}(\mathbf{x})=
\sum_{k\in E}\int_{0}^{T_{\mathrm{sim}}}
\left|Y_{\mathrm{target},k}(t+\delta t_{\mathrm{opt}}) - \big(s_{\mathrm{opt}}\,Y_{\mathrm{sim},k}(t)+r_{\mathrm{opt}}\big)\right|^2\,dt.
\end{equation}

\subsection{Activation-time map analysis}
We repeated the full FCPO calibration 10 ($N_{\mathrm{runs}}$) times with different random seeds (including the LHS initialization). For each run $r$, we extracted the final best solution and computed its activation time field $t_a^{(r)}(\mathbf{x})$ on the ventricular mesh.
\begin{equation}
\sigma_{t_a}(\mathbf{x})=
\sqrt{\frac{1}{N_{\mathrm{runs}}-1}\sum_{r=1}^{N_{\mathrm{runs}}}
\left(t_a^{(r)}(\mathbf{x})-\overline{t_a}(\mathbf{x})\right)^2},
\qquad
\overline{t_a}(\mathbf{x})=
\frac{1}{N_{\mathrm{runs}}}\sum_{r=1}^{N_{\mathrm{runs}}} t_a^{(r)}(\mathbf{x}).
\end{equation}
This quantity provides a spatial measure of identifiability of the recovered activation sequence under repeated optimization.

\section{Results}\label{sec:results}

\begin{figure}[!htbp]
    \centering
    \includegraphics[width=0.9\textwidth]{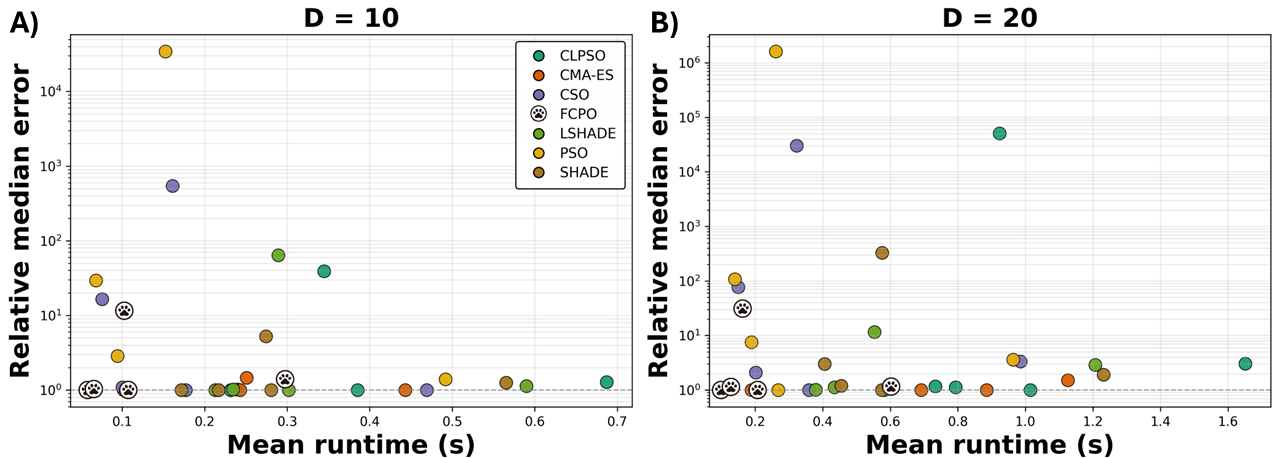}
    % \caption{Accuracy--runtime trade-off at $D=20$ for representative CEC~2022 functions. Each marker shows mean best fitness versus mean runtime across 30 runs; error bars denote one standard deviation in each axis.}
    \caption{Pareto-style runtime--accuracy comparison for the CEC~2022 benchmark subset at \textbf{A)} $D=10$ and \textbf{B)} $D=20$. Markers show each optimizer’s mean runtime and relative median error over 30 runs. Relative median error is normalized by the best median final objective value obtained in each case, so the best method is located at 1 (horizontal dashed line) and larger values indicate worse performance. Lower values on both axes indicate a better trade-off.}
    \label{fig:pareto_d20}
\end{figure}

\begin{figure}[!htbp]
    \centering
    \includegraphics[width=0.9\textwidth]{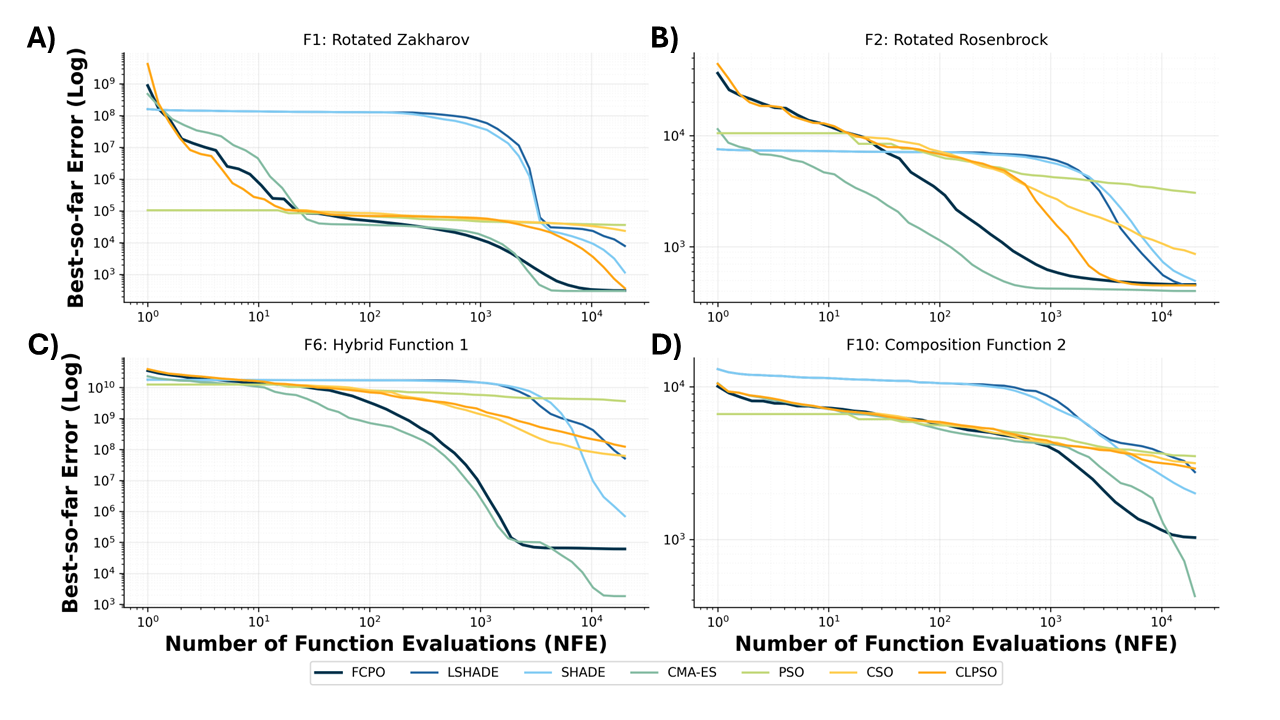}
    % \caption{Best-so-far convergence versus iteration at $D=20$ (mean over 30 runs; shaded band: $\pm$1 std) for representative functions.}
    % \caption{Best-so-far convergence versus the realized number of function evaluations (NFE) at $D=20$ (mean over 30 runs) for representative functions.}
    \caption{Convergence profiles on representative CEC benchmark functions at $D=20$. The best-so-far error is reported as a function of the number of function evaluations (NFE), using logarithmic scales on both axes, for FCPO and six baseline optimizers. Panels correspond to: \textbf{A)} F1: Rotated Zakharov, \textbf{B)} F2: Rotated Rosenbrock, \textbf{C)} F6: Hybrid Function~1, and \textbf{D)} F10: Composition Function~2.}
    \label{fig:convergence_d20}
\end{figure}

Detailed per-function statistics are provided in Supplementary Tables~\ref{tab:benchmark_stats_d10}--\ref{tab:benchmark_stats_d20}. Figure~\ref{fig:pareto_d20} summarizes the accuracy--runtime trade-off at $D=20$, Figure~\ref{fig:convergence_d20} shows best-so-far convergence versus the realized NFE, and Figure~\ref{fig:ablation_panel} isolates the contribution of the main FCPO components on the most challenging benchmark classes. The discussion below emphasizes the dominant qualitative pattern of each function, while the complete numerical results for all methods are left to the Supplementary Tables.

% \section{Results}\label{sec:results}

% Detailed per-function statistics (mean $\pm$ standard deviation of the final best objective and wall-clock runtime over 30 runs) are provided in Supplementary Tables~\ref{tab:benchmark_stats_d10}--\ref{tab:benchmark_stats_d20}. To emphasize the practical accuracy--cost trade-off, Figure~\ref{fig:pareto_d20} summarizes mean objective versus mean runtime at $D=20$, while Figure~\ref{fig:convergence_d20} reports best-so-far convergence versus the realized NFE. Figure~\ref{fig:ablation_panel} isolates the contribution of FCPO components on the hardest benchmark classes.

\FloatBarrier

\subsection{Statistical comparison}
For each function--dimension case, we assessed overall differences among algorithms using a Kruskal--Wallis test on the final best objective values across 30 independent runs. The omnibus test rejected the null hypothesis of equal performance in all 10 benchmark cases, with $p$-values ranging from $1.01\times 10^{-111}$ to $5.68\times 10^{-13}$. Post-hoc comparisons against FCPO were then performed using Dunn's test with Holm correction, and effect sizes were quantified using Cliff's delta.

Under this protocol, FCPO was significantly better than PSO in 9/10 cases and better than CSO in 8/10 cases, typically with large effect sizes. Relative to CLPSO, FCPO was significantly better in 5 cases, significantly worse in 2, and not significantly different in the remaining 3. Against SHADE, FCPO was significantly better in 5 cases and not significantly different in the other 5. Relative to L-SHADE, FCPO was significantly better in 2 cases and statistically indistinguishable in the remaining 8. CMA-ES remained significantly better than FCPO in 8/10 cases, mainly on the structured functions (F1--F3) and on the hybrid function (F6), whereas FCPO was significantly better than CMA-ES on F10 at $D=10$ and statistically indistinguishable on F10 at $D=20$.

At the aggregate level, average ranks computed from the per-case median final objective values again favored CMA-ES. The Friedman test on case-level ranks rejected the null hypothesis of equal overall performance across algorithms (Friedman statistic $=45.02$, $p=9.14\times 10^{-7}$). Among the full methods, CMA-ES achieved the best average rank (2.20), followed by FCPO (4.90) and L-SHADE (5.10), whereas PSO (9.60) and CSO (8.10) ranked last. These results support the interpretation that FCPO offers a robust middle ground between the strongest covariance-based strategy and the weaker classical swarm baselines.

\FloatBarrier

% \subsection{Overall trends across algorithms}
% CMA-ES attains the best mean fitness on the structured functions F1--F3 and on the hybrid function F6, reflecting its robustness under rotation and ill-conditioning. The DE baselines (SHADE and L-SHADE) are competitive on several functions but show sensitivity on the hybrid landscape, especially at higher dimension. Among swarm baselines, CLPSO consistently improves upon standard PSO on the rotated Rosenbrock valley (F2), while CSO is the strongest method on the composition function F10 at $D=10$. FCPO emphasizes a practical accuracy--runtime trade-off: it is consistently faster than CMA-ES across most cases (Table~\ref{tab:agg_rank_time}) and achieves the best mean performance on the multimodal composition function F10 at $D=20$.

\subsection{Overall trends across algorithms}
Across the benchmark set, CMA-ES provides the strongest overall accuracy, particularly on the structured functions F1--F3 and on the hybrid function F6. FCPO, however, consistently improves over the classical swarm baselines and remains competitive with the stronger DE baselines while offering lower runtime in many cases. This trade-off is visible in Figure~\ref{fig:pareto_d20}, where FCPO frequently occupies a favorable low-runtime region while maintaining strong objective values, and in Figure~\ref{fig:convergence_d20}, where FCPO often makes rapid early progress under the matched evaluation budget. The composition function F10 at $D=20$ is the clearest case in which FCPO attains the best mean performance among all tested methods.

% \subsection{Unimodal function (CEC22-F1)}
% On the unimodal Zakharov function (F1) at $D=10$, CMA-ES and both DE variants reach values essentially at the shifted optimum ($\approx 300$), with CMA-ES achieving $3.000\times 10^{2}\pm 0$. FCPO also reaches near-optimal values, attaining $3.000\times 10^{2} \pm 0.00178$ with a lower mean runtime ($0.058 \pm 0.052$\,s vs.\ $0.110 \pm 0.006$\,s for CMA-ES). In contrast, PSO and CSO fail to consistently converge to the optimum basin under the same iteration budget (PSO $9.437\times 10^{3} \pm 3.434\times 10^{3}$; CSO $4.758\times 10^{3} \pm 1.731\times 10^{3}$).

% At $D=20$, CMA-ES remains essentially optimal ($3.000\times 10^{2}\pm 0$), while FCPO achieves $3.073\times 10^{2} \pm 4.561\times 10^{1}$ and strongly improves over PSO and CSO (PSO $3.490\times 10^{4} \pm 9.191\times 10^{3}$; CSO $2.335\times 10^{4} \pm 5.405\times 10^{3}$). CLPSO reduces the error relative to PSO (CLPSO $3.658\times 10^{2} \pm 4.770\times 10^{1}$) but remains less accurate than FCPO/CMA-ES on this unimodal setting.

\subsection{Unimodal function (CEC22-F1)}
On the unimodal Zakharov function (F1), FCPO reaches near-optimal solutions at both dimensions and clearly improves over the classical swarm baselines, although CMA-ES remains the most accurate method overall. At $D=10$, FCPO attains $3.000\times 10^{2} \pm 0.00178$, essentially matching the optimum while remaining faster than CMA-ES ($0.058 \pm 0.052$\,s vs.\ $0.110 \pm 0.006$\,s). At $D=20$, CMA-ES remains optimal, whereas FCPO achieves $3.073\times 10^{2} \pm 4.561\times 10^{1}$ and still substantially outperforms PSO and CSO in solution quality (Supplementary Table~\ref{tab:benchmark_stats_d20}). Overall, F1 shows that FCPO preserves the low-overhead behavior of swarm methods while markedly improving their reliability on smooth unimodal structure, although CMA-ES retains the best absolute accuracy.

% \subsection{Valley-shaped function (CEC22-F2)}
% On the shifted and rotated Rosenbrock function (F2), CMA-ES and CLPSO are particularly strong, consistent with their ability to handle curved valleys under rotation. At $D=10$, CMA-ES achieves $4.008\times 10^{2} \pm 1.622$ and CLPSO follows closely ($4.089\times 10^{2} \pm 1.978$). FCPO attains $4.222\times 10^{2} \pm 2.869\times 10^{1}$ while remaining fast ($0.066 \pm 0.003$\,s) and substantially outperforming PSO ($1.334\times 10^{3} \pm 9.075\times 10^{2}$). The DE baselines occupy an intermediate regime (e.g., SHADE $4.252\times 10^{2} \pm 3.227\times 10^{1}$; L-SHADE $4.454\times 10^{2} \pm 7.051\times 10^{1}$).

% At $D=20$, CMA-ES reaches $4.000\times 10^{2} \pm 0$ and remains the most accurate method. CLPSO and L-SHADE form the next-best group with very low dispersion (CLPSO $4.491\times 10^{2} \pm 0.015$; L-SHADE $4.490\times 10^{2} \pm 0.779$). FCPO achieves $4.566\times 10^{2} \pm 2.318\times 10^{1}$ and is substantially faster than CMA-ES ($0.128 \pm 0.004$\,s vs.\ $0.692 \pm 0.238$\,s).

\subsection{Valley-shaped function (CEC22-F2)}
On the shifted and rotated Rosenbrock function (F2), CMA-ES is the most accurate method at both dimensions, with CLPSO also performing strongly. FCPO remains competitive in final objective value while offering a more favorable runtime profile, as also visible in Figure~\ref{fig:pareto_d20}. At $D=10$, FCPO reaches $4.222\times 10^{2} \pm 2.869\times 10^{1}$, compared with $4.008\times 10^{2} \pm 1.622$ for CMA-ES, but with substantially lower runtime. At $D=20$, FCPO attains $4.566\times 10^{2} \pm 2.318\times 10^{1}$, whereas CMA-ES remains best at $4.000\times 10^{2} \pm 0$ (Supplementary Tables~\ref{tab:benchmark_stats_d10}--\ref{tab:benchmark_stats_d20}). These results indicate that FCPO does not surpass the strongest structured-search methods on curved valley landscapes, but it remains competitive while retaining a substantially better accuracy--runtime trade-off than heavier optimizers.

% \subsection{Expanded Schaffer function (CEC22-F3)}
% On the expanded Schaffer’s F7 function (F3), all methods converge very close to the shifted optimum ($\approx 600$) at both dimensions, leading to small differences in final accuracy. At $D=10$, CMA-ES, FCPO and CSO reach essentially identical best values (CMA-ES $6.000\times 10^{2} \pm 0$; FCPO $6.000\times 10^{2} \pm 0$; CSO $6.000\times 10^{2} \pm 0$), and runtime differences become the dominant practical factor.

% At $D=20$, CMA-ES remains optimal ($6.000\times 10^{2} \pm 0$). FCPO remains near-optimal ($6.000\times 10^{2} \pm 0.0129$) and improves upon PSO ($6.016\times 10^{2} \pm 0.354$) and CSO ($6.002\times 10^{2} \pm 0.059$), while maintaining competitive runtime.

\subsection{Expanded Schaffer function (CEC22-F3)}
On the expanded Schaffer’s F7 function (F3), most methods converge very close to the shifted optimum at both dimensions, so differences in final objective value are small. At $D=10$, FCPO reaches the optimum value ($6.000\times 10^{2} \pm 0$), matching CMA-ES and other strong baselines. At $D=20$, FCPO remains near-optimal with $6.000\times 10^{2} \pm 0.0129$, while maintaining lower runtime than CMA-ES (Supplementary Tables~\ref{tab:benchmark_stats_d10}--\ref{tab:benchmark_stats_d20}). Thus, on F3 the benchmark is effectively saturated by several methods, and the main practical distinction becomes computational cost rather than final accuracy. In this regime, FCPO remains competitive because it achieves near-optimal values without incurring the higher runtime of CMA-ES or the DE baselines.

% \subsection{Hybrid function (CEC22-F6)}
% The hybrid function (F6) is markedly more challenging and reveals clear separations among algorithms. At $D=10$, CMA-ES is the most accurate method ($1.803\times 10^{3} \pm 3.65$). Among non-ES baselines, SHADE provides the next best mean ($9.330\times 10^{3} \pm 2.346\times 10^{3}$), while FCPO attains $2.320\times 10^{4} \pm 1.242\times 10^{4}$ and improves by orders of magnitude over PSO and CSO (PSO $1.011\times 10^{8} \pm 1.020\times 10^{8}$; CSO $1.067\times 10^{6} \pm 7.081\times 10^{5}$). FCPO is also substantially faster than CMA-ES on this task ($0.102 \pm 0.005$\,s vs.\ $0.443 \pm 0.239$\,s).

% At $D=20$, CMA-ES remains clearly best ($1.837\times 10^{3} \pm 2.941\times 10^{1}$) but with higher runtime ($0.885 \pm 0.231$\,s). FCPO achieves $6.120\times 10^{4} \pm 2.337\times 10^{4}$ and is among the fastest methods on this function ($0.162 \pm 0.233$\,s), substantially outperforming PSO, CSO, CLPSO and both DE baselines in mean fitness at this dimension. This indicates that FCPO’s state-conditioned exploration/refinement schedule yields a clear robustness gain relative to classical swarms on the hybrid setting when dimension increases, even though CMA-ES remains superior in absolute accuracy.

\subsection{Hybrid function (CEC22-F6)}
The hybrid function (F6) produces a clearer separation among methods and is one of the most challenging cases in the benchmark. CMA-ES is the best-performing optimizer at both dimensions, while FCPO substantially improves over the classical swarm baselines and remains among the faster methods. At $D=10$, FCPO attains $2.320\times 10^{4} \pm 1.242\times 10^{4}$, far better than PSO and CSO, although still above the CMA-ES result of $1.803\times 10^{3} \pm 3.65$. At $D=20$, FCPO reaches $6.120\times 10^{4} \pm 2.337\times 10^{4}$, whereas CMA-ES remains clearly superior in accuracy (Supplementary Tables~\ref{tab:benchmark_stats_d10}--\ref{tab:benchmark_stats_d20}). The representative convergence curves in Figure~\ref{fig:convergence_d20} are consistent with this behavior, showing that FCPO improves rapidly but does not close the final gap to CMA-ES on this hybrid structure. These results show that FCPO provides a substantial robustness gain over standard swarms on hybrid landscapes, especially as dimensionality increases, but does not yet match the strongest covariance-based strategy on this class of problems.

% \subsection{Composition function (CEC22-F10)}
% On the composition function (F10), FCPO exhibits its clearest advantage as dimension increases. At $D=10$, CSO achieves the best mean ($1.786\times 10^{3} \pm 6.541\times 10^{2}$), followed by L-SHADE ($1.870\times 10^{3} \pm 5.827\times 10^{2}$). FCPO ranks among the better-performing methods ($2.118\times 10^{3} \pm 7.657\times 10^{2}$) and remains faster than CSO and the DE baselines in this configuration (e.g., FCPO $0.297 \pm 0.007$\,s vs.\ CSO $0.469 \pm 0.234$\,s).

% At $D=20$, FCPO achieves the best mean value among all tested methods ($9.625\times 10^{2} \pm 1.275\times 10^{3}$), improving over CMA-ES ($1.505\times 10^{3} \pm 1.717\times 10^{3}$), SHADE ($2.004\times 10^{3} \pm 6.063\times 10^{2}$) and L-SHADE ($2.759\times 10^{3} \pm 9.371\times 10^{2}$). Importantly, FCPO also maintains lower runtime than CMA-ES and the DE baselines (FCPO $0.602 \pm 0.011$\,s vs.\ CMA-ES $1.126 \pm 0.490$\,s and L-SHADE $1.206 \pm 0.321$\,s). The relatively large dispersion on this function indicates a highly rugged landscape, but FCPO’s mean performance and runtime place it on the favorable side of the accuracy--runtime trade-off at $D=20$ (Figure~\ref{fig:pareto_d20}).

\subsection{Composition function (CEC22-F10)}
The composition function (F10) is the case where FCPO exhibits its clearest advantage, particularly at higher dimension. At $D=10$, FCPO is competitive but not the best method, with CSO and L-SHADE achieving lower mean objective values. At $D=20$, however, FCPO attains the best mean objective among all tested methods, $9.625\times 10^{2} \pm 1.275\times 10^{3}$, while also remaining faster than CMA-ES and the DE baselines. In particular, FCPO outperforms CMA-ES in mean objective value at $D=20$ while requiring roughly half the runtime on average, placing it on the favorable side of the accuracy--runtime trade-off in Figure~\ref{fig:pareto_d20}. Figure~\ref{fig:convergence_d20} further shows that FCPO maintains competitive progress as the evaluation budget increases on this rugged landscape. This behavior suggests that FCPO’s combination of state-conditioned exploration and covariance-guided refinement is especially beneficial on multimodal composition structure, where avoiding premature stagnation appears more important than maximizing asymptotic local-search precision.

% \subsection{Computational efficiency}
% Across the ten benchmark cases, FCPO achieves the lowest average runtime among the full methods (mean $0.183$\,s; Table~\ref{tab:agg_rank_time}), while remaining the second-best family by average rank behind CMA-ES. Disabling population size reduction (FCPO\_NoLPSR) improves average rank but increases average runtime (Table~\ref{tab:agg_rank_time}), highlighting the intended accuracy--runtime trade-off controlled by LPSR. %Runtime variability across trials for representative functions at $D=20$ is summarized in Figure~\ref{fig:runtime_box_d20}.

\subsection{Computational efficiency}
Across the 10 benchmark cases, FCPO achieves the lowest mean runtime among the full methods (Table~\ref{tab:agg_rank_time}), while remaining competitive in average rank. This pattern is consistent with the accuracy--runtime trade-off shown in Figure~\ref{fig:pareto_d20}, where FCPO frequently lies in the favorable low-runtime region without a corresponding collapse in solution quality. Disabling population size reduction (\texttt{FCPO\_NoLPSR}) improves the aggregate rank but increases runtime, confirming that LPSR is a key contributor to the practical efficiency of the full FCPO configuration.

% \subsection{Ablation analysis of FCPO}\label{subsec:ablation}
% To isolate the contributions of FCPO’s components, we evaluate three ablations: FCPO\_NoZoom (removing the Zoomies jump operator), FCPO\_NoEigen (removing eigen-aligned local refinement), and FCPO\_NoLPSR (disabling population size reduction). On the easier functions (F1 and F3), all variants converge near the optimum and differences are negligible and depicted in Figure~\ref{fig:ablation_panel}.

\subsection{Ablation analysis of FCPO}\label{subsec:ablation}
\begin{figure}[!htbp]
    \centering
    \includegraphics[width=0.9\textwidth]{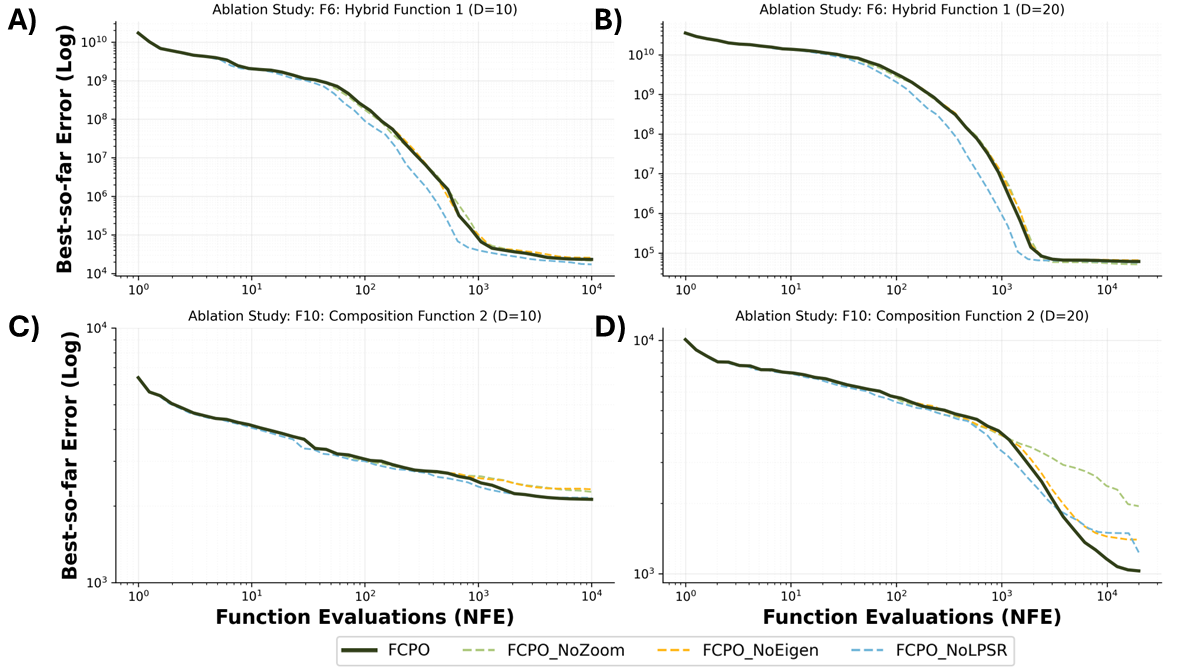}
    % \caption{Ablation study of FCPO components (Zoomies, eigen-refinement, LPSR) on representative benchmark cases.}
    \caption{Ablation analysis of FCPO on representative benchmark functions. The convergence behavior of the full FCPO algorithm is compared with three reduced variants, namely FCPO\_NoZoom, FCPO\_NoEigen, and FCPO\_NoLPSR, using NFE-matched best-so-far error curves. Panels correspond to: \textbf{A)} F6: Hybrid Function~1 ($D=10$), \textbf{B)} F6: Hybrid Function~1 ($D=20$), \textbf{C)} F10: Composition Function~2 ($D=10$), and \textbf{D)} F10: Composition Function~2 ($D=20$).}
    \label{fig:ablation_panel}
\end{figure}

To isolate the contributions of the main FCPO components, we evaluate three ablations: \texttt{FCPO\_NoZoom} (removing the Zoomies jump operator), \texttt{FCPO\_NoEigen} (removing eigen-aligned local refinement), and \texttt{FCPO\_NoLPSR} (disabling population size reduction). Figure~\ref{fig:ablation_panel} summarizes the resulting trends on representative difficult cases, while the complete numerical values are provided in Supplementary Tables~\ref{tab:benchmark_stats_d10}--\ref{tab:benchmark_stats_d20}. On the easier functions (F1 and F3), all variants converge near the optimum and the differences are negligible, indicating that the ablated components mainly matter on rugged or hybrid landscapes.

\FloatBarrier

On the more difficult landscapes, the ablations reveal function-dependent trade-offs. On F10 at $D=20$, removing Zoomies deteriorates performance from $9.625\times 10^{2} \pm 1.275\times 10^{3}$ to $1.757\times 10^{3} \pm 9.536\times 10^{2}$, while removing eigen refinement yields $1.329\times 10^{3} \pm 1.451\times 10^{3}$. Disabling LPSR increases runtime substantially (from $0.602 \pm 0.011$\,s to $0.976 \pm 0.243$\,s) and does not improve the mean objective on this composition landscape ($1.233\times 10^{3} \pm 1.267\times 10^{3}$). In contrast, on the hybrid function F6 at $D=20$ the no-Zoomies and no-LPSR variants slightly improve the mean objective ($5.155\times 10^{4} \pm 2.290\times 10^{4}$ and $5.173\times 10^{4} \pm 2.017\times 10^{4}$ vs.\ $6.120\times 10^{4} \pm 2.337\times 10^{4}$), suggesting that long-range jumps and aggressive population reduction are most beneficial on rugged composition structure, whereas the hybrid function benefits more from stable exploitation once a reasonable basin is reached. Overall, the ablations support that FCPO’s strongest results on composition landscapes arise from combining (i) episodic global jumps and (ii) covariance-informed eigen-aligned refinement, while LPSR primarily governs the runtime advantage.

\subsection{Ventricular activation digital-twin calibration}
\begin{figure}[!htb]
    \centering
    \includegraphics[width=0.9\textwidth]{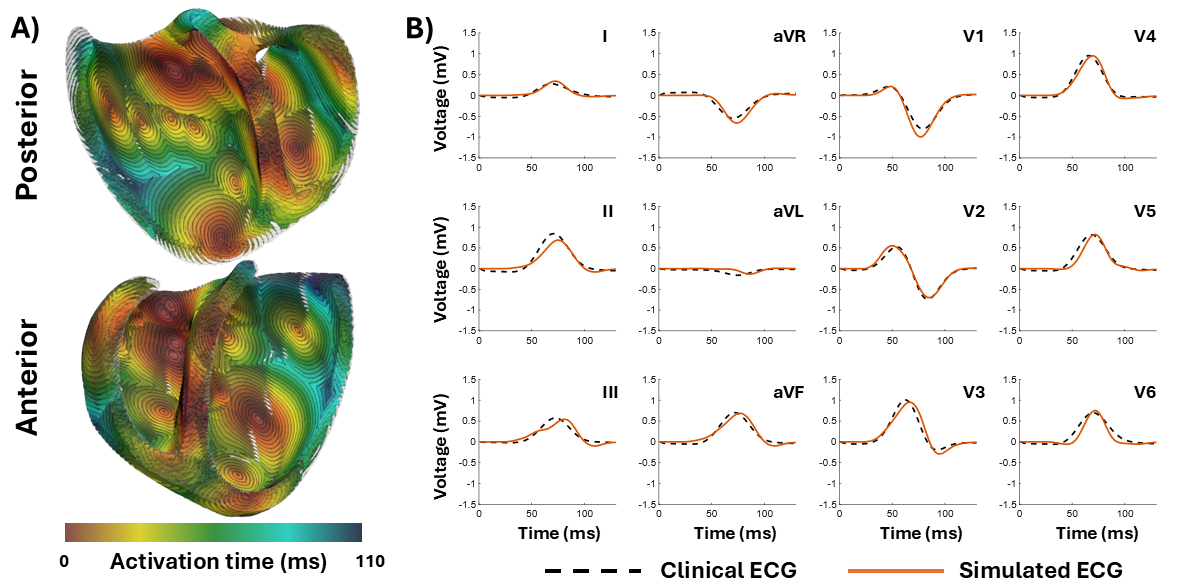}
    \caption{Ventricular activation digital-twin calibration using FCPO.
    \textbf{A)} Simulated biventricular activation time map (posterior and anterior views) obtained from the optimized Purkinje--myocyte junctions (PMJs) distribution; colors denote local activation time (ms) relative to the onset, with the range shown in the color bar.
    \textbf{B)} Twelve-lead ECG comparison between the clinical target (black dashed) and the ECG simulated from the optimized activation pattern (orange solid).}
    \label{fig:dt_ecg}
\end{figure}

In the application to cardiac digital twinning, FCPO successfully converged to a solution (RMSE $<0.1$\,mV) within approximately 40 iterations. Figure~\ref{fig:dt_ecg}A shows the resulting biventricular activation time field produced by the optimized PMJ configuration, visualized on the ventricular geometry from posterior and anterior viewpoints. Figure~\ref{fig:dt_ecg}B compares the target clinical ECG (black dashed) with the ECG simulated from the optimized activation pattern (orange solid) across all 12 leads.

\FloatBarrier

Qualitatively, the optimized digital twin reproduces the main depolarization features of the clinical ECG across all 12 leads, including the lead-dependent polarity pattern and the overall QRS morphology. The reconstructed ventricular activation sequence spans approximately \(0\) to \(110\) ms over the ventricular surfaces (Figure~\ref{fig:dt_ecg}A), and the simulated ECG traces show close agreement with the clinical recordings in both polarity and waveform shape (Figure~\ref{fig:dt_ecg}B). The temporal alignment of the depolarization phase is close, with only small residual peak mismatches, indicating that FCPO successfully tuned both the onset delays \(t_{\text{onset},i}\) and the spatial PMJ coordinates to recover a patient-specific activation sequence. In comparative trials under the same forward-model configuration and a 100-iteration budget, CMA-ES achieved similar ECG fidelity at the cost of a 2.3-fold increase in computational time, whereas L-SHADE did not converge to the target.

\begin{figure}[!htb]
    \centering
    \includegraphics[width=0.9\textwidth]{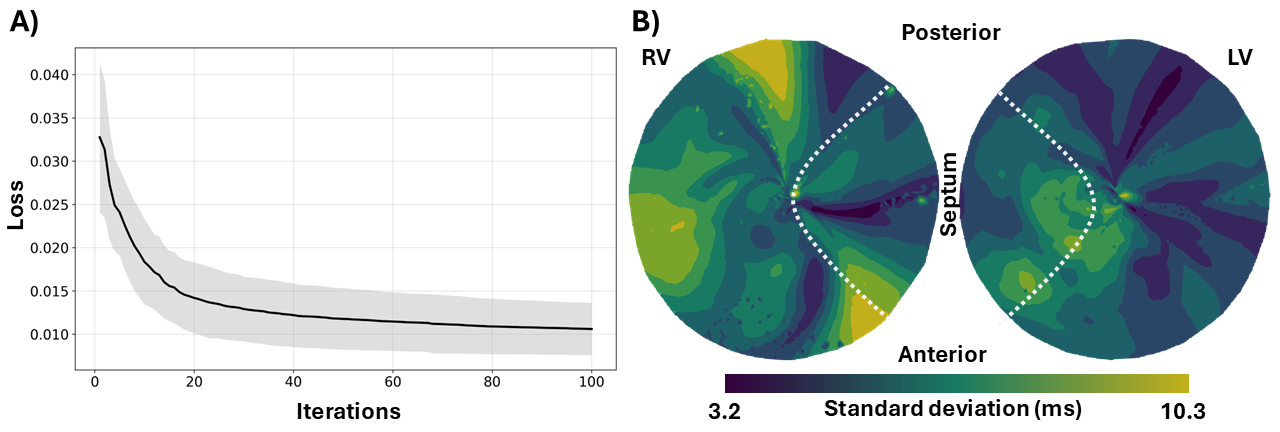}
    \caption{Convergence and activation-time standard deviation of FCPO in ventricular digital-twin calibration. \textbf{A)} Best loss versus iteration, shown as the mean over 10 independent runs (black line) with $\pm$1 standard deviation (gray band). \textbf{B)} Spatial endocardial activation variability over 10 independent runs, shown as the nodewise standard deviation of activation time, $\sigma_{t_a}(\mathbf{x})$, computed across all final best solutions from the repeated runs and mapped onto the ventricular mesh. Low $\sigma_{t_a}$ indicates anatomically identifiable regions, whereas high $\sigma_{t_a}$ highlights ill-posed areas where multiple activation patterns yield comparable ECG fit.}
    \label{fig:act_std}
\end{figure}

Figure~\ref{fig:act_std}A summarizes the convergence behavior of FCPO over 10 independent calibrations with different random initializations. The mean loss decreases rapidly from approximately \(3.3\times 10^{-2}\) at the first iteration to about \(1.5\times 10^{-2}\) within the first 15--20 iterations, reaches the target fidelity threshold of RMSE \(< 0.1\) mV at around 40 iterations, and then plateaus near \(1.0\times 10^{-2}\) to \(1.1\times 10^{-2}\) by iteration 100. Figure~\ref{fig:act_std}B complements this analysis by showing the nodewise standard deviation of activation time, \(\sigma_{t_a}(\mathbf{x})\), across the final best solutions from the 10 runs. The spatial variability ranges from \(3.2\) to \(10.3\)\,ms, indicating that some regions are reconstructed more consistently than others. Regions with low \(\sigma_{t_a}\) are better constrained by the ECG data, whereas regions with higher \(\sigma_{t_a}\) indicate reduced identifiability, meaning that multiple activation patterns can produce similarly good ECG fits. Together, these results show that FCPO can recover physiologically plausible activation patterns under a limited iteration budget while also revealing the spatial distribution of reconstruction uncertainty.

\FloatBarrier

\section{Discussion}

FCPO was designed as a lightweight hybrid swarm optimizer for limited-budget expensive black-box optimization, combining a PSO backbone with particle-wise Markov state switching, episodic elite-difference exploration, covariance-guided local refinement, and linear population size reduction. The benchmark results indicate that this combination yields a favorable accuracy--runtime trade-off: across the tested cases, FCPO consistently improves over the classical swarm baselines, remains competitive with the stronger DE variants, and achieves the lowest mean runtime among the full methods, while CMA-ES remains the strongest method in overall accuracy. Statistically, this picture is supported both by the per-case nonparametric comparisons and by the aggregate rank analysis, which place FCPO between the strongest covariance-based strategy and the weaker classical swarms.

\subsection{FCPO as adaptive operator scheduling rather than a metaphor}
Although FCPO is motivated by feline ethology, its methodological contribution is best understood as an adaptive operator-scheduling framework rather than as a metaphor-driven heuristic. In this view, the key idea is that different search operators should be activated at different moments of the search, depending on the current balance between exploitation, restoration, and exploration. FCPO implements this through a particle-wise Markov controller that coordinates a PSO-style update, an elite-difference exploration step, a covariance-guided local refinement step, and a damped restoration move. This interpretation is more informative than the metaphor itself and aligns the method with the broader literature on hybrid metaheuristics and adaptive operator selection~\cite{Sorensen2015,Fialho2010,Maturana2012}.

\subsection{Role of Markov switching and state-dependent search}
A central feature of FCPO is that the swarm does not follow a single homogeneous update rule throughout the run. Instead, each particle carries an explicit discrete state, and this state determines which operator is applied at a given iteration. The Markov transition matrix is adapted online by reinforcing transitions toward the state of the current best particle and by increasing the probability of exploration under stagnation. This gives the method a simple but flexible mechanism for coordinating heterogeneous behaviors without requiring additional function evaluations. Conceptually, this differs from deterministic phase schedules and from globally shared mode-switching rules, because particles may follow different stochastic trajectories even when they are in the same region of the search space. The empirical results suggest that this state-conditioned structure is beneficial primarily on difficult multimodal landscapes rather than on smooth structured ones. On the structured cases, such as F1--F3, the advantage of adaptive switching is limited because several strong methods already reach near-optimal solutions. On the more rugged landscapes, especially F10 at $D=20$, FCPO appears to benefit from being able to alternate between rapid swarm-style progress and disruptive exploration, while still recovering local directional information from elites.

\subsection{When FCPO helps most: rugged multimodal and composition landscapes}
The clearest empirical advantage of FCPO appears on the composition function F10 at $D=20$, where it achieves the best mean final objective value among all tested methods while remaining faster than CMA-ES and the DE baselines. This is also the regime in which the ablation study is most informative: removing Zoomies or eigen-guided refinement degrades performance, whereas disabling LPSR increases runtime without improving the mean objective. Taken together, these observations support the interpretation that FCPO is particularly well suited to rugged multimodal landscapes in which premature stagnation is a major risk and where a purely local covariance-driven search may be insufficient on its own. In contrast, on the hybrid function F6 and on the more structured cases, CMA-ES remains clearly stronger in final accuracy. This indicates that FCPO should not be interpreted as a universal replacement for strong covariance-based evolution strategies. Rather, the results suggest that its main advantage lies in providing a practical balance between exploration capacity, low implementation overhead, and acceptable late-stage refinement under limited budgets.

\subsection{Accuracy--runtime trade-off under matched evaluation budgets}
A key point of the present study is that all methods were compared under a matched maximum function-evaluation budget, and convergence was analyzed versus realized NFE. This makes the comparisons more meaningful in the context of expensive objectives, where the dominant cost is often the repeated forward evaluation rather than the optimizer update itself. Within this setting, FCPO achieves the lowest mean runtime among the full methods while remaining competitive in average rank. The aggregate analysis therefore supports a practical interpretation of FCPO as a low-latency optimizer that sacrifices some asymptotic accuracy relative to CMA-ES in exchange for more favorable runtime behavior and stronger robustness than the classical swarm baselines. This trade-off is also visible qualitatively in the Pareto-style plots at $D=20$, where FCPO repeatedly appears in the favorable low-runtime region without a corresponding collapse in solution quality. In other words, the method does not dominate the benchmark set in accuracy, but it often occupies a useful operating regime for practitioners who need strong solutions quickly under a fixed evaluation budget.

\subsection{Relevance to cardiac digital twinning}

Although this paper focuses on optimization, the motivating use case (ECG-driven model personalization) is representative of the intended regime: expensive, ill-posed inverse problems with nonconvex objectives and limited evaluation budgets. Prior work by \cite{Gillette2021} has shown that ECG simulation and personalization pipelines can be accelerated using lead-field formulations~\citep{Potse2018} and simplified propagation models~\citep{Neic2017,BarriosEspinosa2025}, enabling iterative calibration workflows from surface ECG data~\citep{Camps2021}. Related ECG-based personalization and activation reconstruction studies further underline the need for robust optimizers under strong identifiability challenges and heterogeneous cost landscapes~\citep{Pezzuto2022,Grandits2020InverseEikonal,Grandits2022Smoothness,Grandits2025}.

A key point emerging from these studies is that the dominant cost driver is often the number of forward solves, not the optimizer’s per-iteration overhead. For example, \cite{Camps2021} report Bayesian calibration workflows that may require on the order of $10^4$ forward simulations for a single case, with end-to-end runtimes on the order of tens of hours depending on the chosen propagation surrogate; they also highlight how per-simulation cost can vary widely across model variants, making NFE the critical budget in practice~\citep{Camps2021,Camps2024PurkinjeTwin,Camps2025Repol}. In parallel, \cite{Gillette2021} demonstrate that even when anatomical steps are tractable (reporting an anatomical twinning stage on the order of hours for a small cohort), functional personalization still benefits strongly from accelerations such as lead-field reuse and simplified propagation models, precisely because calibration is iterative and simulation-limited. Finally, \cite{Grandits2021GEASI} (GEASI) show that exploiting problem structure (geodesics and Gauss--Newton-type updates) can produce convincing fits for activation-time targets within $\mathcal{O}(10^2)$ iterations, while ECG-driven 3D optimizations can remain multi-hour even with GPU-parallelized inner loops, emphasizing the sensitivity of overall runtime to the repeated inner computations required by the forward ECG~\citep{Grandits2022Smoothness,Grandits2025}.

In this context, the practical relevance of FCPO is its explicit design around low-latency search under tight evaluation budgets: LPSR~\citep{Tanabe2014} reduces evaluations as the run progresses, and the Markov switching controller concentrates ``high-energy'' exploration (Zoomies) only when stagnation is detected, while keeping the baseline dynamics lightweight (PSO-like)~\citep{Kennedy1995,Chu2006}. Compared to inference/sampling-heavy personalization strategies (e.g., Bayesian or ABC-style calibration~\citep{Camps2021}) FCPO targets the regime where a user must obtain a competitive solution in hundreds to low-thousands of forward evaluations, rather than tens of thousands. Compared to structure-exploiting approaches like GEASI and related inverse-eikonal methods~\citep{Grandits2021GEASI,Grandits2020InverseEikonal}, FCPO remains fully derivative-free and agnostic to the internals of the objective (including non-smooth misfit terms or simulator switches), making it a complementary option when differentiability/adjoints are unavailable or unreliable. Overall, the comparisons suggest that, for ECG-driven personalization pipelines where identifiability is limited and each evaluation can be seconds to minutes, reducing NFE while retaining reliable escape mechanisms from stagnation can be more impactful than marginal asymptotic gains in best-known objective value.

\subsection{Limitations}
Several limitations should be emphasized. First, although the experiments were carried out under a matched function-evaluation budget and the realized NFE is reported for transparency, the benchmark remains limited in scope. Only five representative CEC~2022 functions and two dimensional settings were used, so the conclusions should be interpreted as evidence on a focused subset rather than as a comprehensive benchmark of global optimizer performance. Broader budget sweeps, larger function sets, and additional benchmark families would provide a more complete picture~\citep{Awad2016,Kumar2021,Latorre2021,Piotrowski2025}. Second, FCPO combines several interacting ingredients, and the present ablation study does not yet isolate the Markov controller as sharply as it could. While the current ablations clarify the roles of Zoomies, eigen-guided refinement, and LPSR, future work should test variants with fixed transition matrices, simplified neutral-state structures, or no adaptive reinforcement in order to quantify more precisely how much of the observed gain comes from the state-switching controller itself. Third, the cardiac study is still illustrative rather than fully validating. Additional experiments across multiple subjects, initialization regimes, and forward-model fidelities will be needed to determine more reliably where FCPO provides the greatest practical benefit relative to structure-exploiting or gradient-based calibration methods. Such studies should also assess success rates, uncertainty under noise and model mismatch, and alternative ECG loss functions that may better reflect morphology-aware agreement.

\section{Conclusion}

We presented \emph{Frenetic Cat-inspired Particle Optimization (FCPO)}, a hybrid swarm optimizer that formalizes stochastic, particle-wise switching between exploration and refinement operators via an explicit-state Markov controller. FCPO combines a lightweight PSO backbone with episodic DE-style elite-difference jumps (Zoomies), covariance-informed eigen-space local refinement (Purr), and linear population size reduction to control late-stage computational effort.

On five representative CEC~2022 benchmark functions at $D\in\{10,20\}$, FCPO achieves the lowest mean runtime across the tested methods and is the best-performing method in mean objective value on the challenging composition function F10 at $D=20$. CMA-ES remains the most accurate optimizer on the smooth/structured functions and on the hybrid function, highlighting that FCPO is best viewed as a practical accuracy--runtime trade-off method for limited-budget, expensive objectives rather than a universal replacement for strong evolutionary strategies. Finally, FCPO successfully performs ventricular activation digital-twin calibration from 12-lead ECGs within $\sim 40$ iterations, producing physiologically plausible activation maps and robust convergence across repeated initializations.

\section*{Acknowledgments}
This work was supported by the Direcci\'on General de Pol\'itica Cient\'ifica de la Generalitat Valenciana (CIPROM/2023/14) and by PID2022-140553OBC41 [MICIU/AEI/10.13039/501100011033 and by ERDF/EU]. Additionally, funding for open access charge: CRUE--Universitat Polit\`ecnica de Val\`encia.

\section*{Declaration of competing interest}
The authors declare that they have no known competing financial interests or personal relationships that could have appeared to influence the work reported in this paper.

\section*{Data availability}
The code and benchmark data supporting the findings of this study are publicly available at \url{https://github.com/jorge221/fcpo}. The clinical ECG data used in the cardiac digital twin use case are not publicly available due to privacy and ethical restrictions.

% =========================================================
% BIBLIOGRAPHY
% =========================================================
\bibliographystyle{plainnat}
\bibliography{cas-refs}

% =========================================================
% APPENDIX / SUPPLEMENTARY MATERIAL
% =========================================================
\clearpage
\appendix
\FloatBarrier

\setcounter{table}{0}
\renewcommand{\thetable}{S\arabic{table}}

\setcounter{figure}{0}
\renewcommand{\thefigure}{S\arabic{figure}}

\setcounter{equation}{0}
\renewcommand{\theequation}{S\arabic{equation}}

\section*{Supplementary material}

\begin{table}[!htpb]
\centering
\scriptsize
\setlength{\tabcolsep}{6pt}
\caption{CEC~2022 results at $D=10$ (mean $\pm$ standard deviation over 30 independent runs). Final best objective values are reported for each method; runtimes are given in seconds.}
\label{tab:benchmark_stats_d10}
\begin{tabular}{llcc}
\toprule
Function & Algorithm & Final best objective value & Time (s) \\
\midrule
\multirow{10}{*}{F1} & PSO & $9.437\times 10^{3} \pm 3.434\times 10^{3}$ & $0.068 \pm 0.003$\,s \\
 & CSO & $4.758\times 10^{3} \pm 1.731\times 10^{3}$ & $0.200 \pm 0.006$\,s \\
 & CLPSO & $4.557\times 10^{2} \pm 5.206\times 10^{1}$ & $0.159 \pm 0.004$\,s \\
 & SHADE & $3.000\times 10^{2} \pm 0$ & $0.324 \pm 0.007$\,s \\
 & L-SHADE & $\textbf{$3.000\times 10^{2} \pm 0$}$ & $0.213 \pm 0.008$\,s \\
 & CMA-ES & $3.000\times 10^{2} \pm 0$ & $0.110 \pm 0.006$\,s \\
 & FCPO & $3.000\times 10^{2} \pm 0.00178$ & $0.058 \pm 0.052$\,s \\
 & FCPO\_NoZoom & $3.000\times 10^{2} \pm 0.00173$ & $0.057 \pm 0.008$\,s \\
 & FCPO\_NoEigen & $3.000\times 10^{2} \pm 0.00381$ & $0.058 \pm 0.011$\,s \\
 & FCPO\_NoLPSR & $3.000\times 10^{2} \pm 0.00137$ & $0.060 \pm 0.011$\,s \\
\midrule
\multirow{10}{*}{F2} & PSO & $1.334\times 10^{3} \pm 9.075\times 10^{2}$ & $0.075 \pm 0.004$\,s \\
 & CSO & $4.480\times 10^{2} \pm 3.294\times 10^{1}$ & $0.225 \pm 0.006$\,s \\
 & CLPSO & $4.089\times 10^{2} \pm 1.978$ & $0.164 \pm 0.004$\,s \\
 & SHADE & $4.252\times 10^{2} \pm 3.227\times 10^{1}$ & $0.334 \pm 0.006$\,s \\
 & L-SHADE & $4.454\times 10^{2} \pm 7.051\times 10^{1}$ & $0.218 \pm 0.005$\,s \\
 & CMA-ES & $\textbf{$4.008\times 10^{2} \pm 1.622$}$ & $0.260 \pm 0.154$\,s \\
 & FCPO & $4.222\times 10^{2} \pm 2.869\times 10^{1}$ & $0.066 \pm 0.003$\,s \\
 & FCPO\_NoZoom & $4.331\times 10^{2} \pm 5.358\times 10^{1}$ & $0.065 \pm 0.004$\,s \\
 & FCPO\_NoEigen & $4.232\times 10^{2} \pm 3.357\times 10^{1}$ & $0.067 \pm 0.004$\,s \\
 & FCPO\_NoLPSR & $4.262\times 10^{2} \pm 3.424\times 10^{1}$ & $0.058 \pm 0.231$\,s \\
\midrule
\multirow{10}{*}{F3} & PSO & $6.004\times 10^{2} \pm 0.146$ & $0.075 \pm 0.004$\,s \\
 & CSO & $6.000\times 10^{2} \pm 0$ & $0.248 \pm 0.007$\,s \\
 & CLPSO & $6.000\times 10^{2} \pm 0$ & $0.181 \pm 0.004$\,s \\
 & SHADE & $6.000\times 10^{2} \pm 0$ & $0.371 \pm 0.007$\,s \\
 & L-SHADE & $6.000\times 10^{2} \pm 0$ & $0.241 \pm 0.008$\,s \\
 & CMA-ES & $\textbf{$6.000\times 10^{2} \pm 0$}$ & $0.102 \pm 0.006$\,s \\
 & FCPO & $6.000\times 10^{2} \pm 0$ & $0.107 \pm 0.004$\,s \\
 & FCPO\_NoZoom & $6.000\times 10^{2} \pm 0$ & $0.106 \pm 0.008$\,s \\
 & FCPO\_NoEigen & $6.000\times 10^{2} \pm 0$ & $0.108 \pm 0.005$\,s \\
 & FCPO\_NoLPSR & $6.000\times 10^{2} \pm 0$ & $0.110 \pm 0.006$\,s \\
\midrule
\multirow{10}{*}{F6} & PSO & $1.011\times 10^{8} \pm 1.020\times 10^{8}$ & $0.093 \pm 0.005$\,s \\
 & CSO & $1.067\times 10^{6} \pm 7.081\times 10^{5}$ & $0.271 \pm 0.008$\,s \\
 & CLPSO & $1.106\times 10^{7} \pm 6.332\times 10^{6}$ & $0.222 \pm 0.006$\,s \\
 & SHADE & $9.330\times 10^{3} \pm 2.346\times 10^{3}$ & $0.363 \pm 0.006$\,s \\
 & L-SHADE & $1.559\times 10^{7} \pm 2.393\times 10^{7}$ & $0.237 \pm 0.006$\,s \\
 & CMA-ES & $\textbf{$1.803\times 10^{3} \pm 3.65$}$ & $0.443 \pm 0.239$\,s \\
 & FCPO & $2.320\times 10^{4} \pm 1.242\times 10^{4}$ & $0.102 \pm 0.005$\,s \\
 & FCPO\_NoZoom & $2.327\times 10^{4} \pm 1.211\times 10^{4}$ & $0.102 \pm 0.007$\,s \\
 & FCPO\_NoEigen & $2.517\times 10^{4} \pm 1.225\times 10^{4}$ & $0.111 \pm 0.011$\,s \\
 & FCPO\_NoLPSR & $1.725\times 10^{4} \pm 1.107\times 10^{4}$ & $0.108 \pm 0.011$\,s \\
\midrule
\multirow{10}{*}{F10} & PSO & $2.504\times 10^{3} \pm 8.668\times 10^{2}$ & $0.491 \pm 0.235$\,s \\
 & CSO & $\textbf{$1.786\times 10^{3} \pm 6.541\times 10^{2}$}$ & $0.469 \pm 0.234$\,s \\
 & CLPSO & $2.197\times 10^{3} \pm 2.638\times 10^{2}$ & $0.347 \pm 0.004$\,s \\
 & SHADE & $2.048\times 10^{3} \pm 5.574\times 10^{2}$ & $0.776 \pm 0.244$\,s \\
 & L-SHADE & $1.870\times 10^{3} \pm 5.827\times 10^{2}$ & $0.590 \pm 0.234$\,s \\
 & CMA-ES & $2.710\times 10^{3} \pm 4.658\times 10^{1}$ & $0.251 \pm 0.005$\,s \\
 & FCPO & $2.118\times 10^{3} \pm 7.657\times 10^{2}$ & $0.297 \pm 0.007$\,s \\
 & FCPO\_NoZoom & $2.474\times 10^{3} \pm 8.433\times 10^{2}$ & $0.281 \pm 0.006$\,s \\
 & FCPO\_NoEigen & $2.223\times 10^{3} \pm 7.091\times 10^{2}$ & $0.304 \pm 0.007$\,s \\
 & FCPO\_NoLPSR & $2.318\times 10^{3} \pm 8.500\times 10^{2}$ & $0.485 \pm 0.237$\,s \\
\bottomrule
\end{tabular}
\end{table}

\begin{table}[!t]
\centering
\scriptsize
\setlength{\tabcolsep}{6pt}
\caption{CEC~2022 results at $D=20$ (mean $\pm$ standard deviation over 30 independent runs). Final best objective values are reported for each method; runtimes are given in seconds.}
\label{tab:benchmark_stats_d20}
\begin{tabular}{llcc}
\toprule
Function & Algorithm & Final best objective value & Time (s) \\
\midrule
\multirow{10}{*}{F1} & PSO & $3.490\times 10^{4} \pm 9.191\times 10^{3}$ & $0.093 \pm 0.005$\,s \\
 & CSO & $2.335\times 10^{4} \pm 5.405\times 10^{3}$ & $0.324 \pm 0.007$\,s \\
 & CLPSO & $3.658\times 10^{2} \pm 4.770\times 10^{1}$ & $0.233 \pm 0.005$\,s \\
 & SHADE & $4.334\times 10^{2} \pm 1.081\times 10^{2}$ & $0.476 \pm 0.008$\,s \\
 & L-SHADE & $7.850\times 10^{3} \pm 1.203\times 10^{4}$ & $0.309 \pm 0.007$\,s \\
 & CMA-ES & $\textbf{$3.000\times 10^{2} \pm 0$}$ & $0.246 \pm 0.163$\,s \\
 & FCPO & $3.073\times 10^{2} \pm 4.561\times 10^{1}$ & $0.098 \pm 0.004$\,s \\
 & FCPO\_NoZoom & $3.153\times 10^{2} \pm 5.087\times 10^{1}$ & $0.097 \pm 0.005$\,s \\
 & FCPO\_NoEigen & $3.612\times 10^{2} \pm 1.779\times 10^{2}$ & $0.099 \pm 0.004$\,s \\
 & FCPO\_NoLPSR & $3.555\times 10^{2} \pm 1.032\times 10^{2}$ & $0.110 \pm 0.003$\,s \\
\midrule
\multirow{10}{*}{F2} & PSO & $3.034\times 10^{3} \pm 1.003\times 10^{3}$ & $0.104 \pm 0.004$\,s \\
 & CSO & $8.626\times 10^{2} \pm 2.000\times 10^{2}$ & $0.379 \pm 0.006$\,s \\
 & CLPSO & $4.491\times 10^{2} \pm 0.015$ & $0.269 \pm 0.004$\,s \\
 & SHADE & $4.599\times 10^{2} \pm 2.237\times 10^{1}$ & $0.547 \pm 0.007$\,s \\
 & L-SHADE & $4.490\times 10^{2} \pm 0.779$ & $0.362 \pm 0.006$\,s \\
 & CMA-ES & $\textbf{$4.000\times 10^{2} \pm 0$}$ & $0.692 \pm 0.238$\,s \\
 & FCPO & $4.566\times 10^{2} \pm 2.318\times 10^{1}$ & $0.128 \pm 0.004$\,s \\
 & FCPO\_NoZoom & $4.535\times 10^{2} \pm 1.141\times 10^{1}$ & $0.129 \pm 0.004$\,s \\
 & FCPO\_NoEigen & $4.518\times 10^{2} \pm 8.287$ & $0.129 \pm 0.004$\,s \\
 & FCPO\_NoLPSR & $4.479\times 10^{2} \pm 0.722$ & $0.244 \pm 0.007$\,s \\
\midrule
\multirow{10}{*}{F3} & PSO & $6.016\times 10^{2} \pm 0.354$ & $0.104 \pm 0.004$\,s \\
 & CSO & $6.002\times 10^{2} \pm 0.059$ & $0.421 \pm 0.007$\,s \\
 & CLPSO & $6.001\times 10^{2} \pm 0.019$ & $0.295 \pm 0.004$\,s \\
 & SHADE & $6.002\times 10^{2} \pm 0.019$ & $0.605 \pm 0.007$\,s \\
 & L-SHADE & $6.002\times 10^{2} \pm 0.037$ & $0.395 \pm 0.006$\,s \\
 & CMA-ES & $\textbf{$6.000\times 10^{2} \pm 0$}$ & $0.463 \pm 0.237$\,s \\
 & FCPO & $6.000\times 10^{2} \pm 0.0129$ & $0.160 \pm 0.004$\,s \\
 & FCPO\_NoZoom & $6.000\times 10^{2} \pm 0.00993$ & $0.160 \pm 0.004$\,s \\
 & FCPO\_NoEigen & $6.000\times 10^{2} \pm 0.0212$ & $0.159 \pm 0.004$\,s \\
 & FCPO\_NoLPSR & $6.000\times 10^{2} \pm 0.0191$ & $0.287 \pm 0.008$\,s \\
\midrule
\multirow{10}{*}{F6} & PSO & $3.389\times 10^{9} \pm 1.857\times 10^{9}$ & $0.132 \pm 0.004$\,s \\
 & CSO & $6.203\times 10^{7} \pm 2.695\times 10^{7}$ & $0.541 \pm 0.007$\,s \\
 & CLPSO & $1.233\times 10^{8} \pm 8.236\times 10^{7}$ & $0.352 \pm 0.005$\,s \\
 & SHADE & $7.015\times 10^{5} \pm 3.394\times 10^{5}$ & $0.681 \pm 0.007$\,s \\
 & L-SHADE & $5.137\times 10^{7} \pm 1.292\times 10^{8}$ & $0.450 \pm 0.006$\,s \\
 & CMA-ES & $\textbf{$1.837\times 10^{3} \pm 2.941\times 10^{1}$}$ & $0.885 \pm 0.231$\,s \\
 & FCPO & $6.120\times 10^{4} \pm 2.337\times 10^{4}$ & $0.162 \pm 0.233$\,s \\
 & FCPO\_NoZoom & $5.155\times 10^{4} \pm 2.290\times 10^{4}$ & $0.161 \pm 0.007$\,s \\
 & FCPO\_NoEigen & $6.593\times 10^{4} \pm 2.433\times 10^{4}$ & $0.170 \pm 0.007$\,s \\
 & FCPO\_NoLPSR & $5.173\times 10^{4} \pm 2.017\times 10^{4}$ & $0.272 \pm 0.241$\,s \\
\midrule
\multirow{10}{*}{F10} & PSO & $3.451\times 10^{3} \pm 5.609\times 10^{2}$ & $0.976 \pm 0.242$\,s \\
 & CSO & $3.157\times 10^{3} \pm 7.193\times 10^{2}$ & $1.039 \pm 0.232$\,s \\
 & CLPSO & $2.911\times 10^{3} \pm 4.051\times 10^{2}$ & $0.734 \pm 0.007$\,s \\
 & SHADE & $2.004\times 10^{3} \pm 6.063\times 10^{2}$ & $1.582 \pm 0.313$\,s \\
 & L-SHADE & $2.759\times 10^{3} \pm 9.371\times 10^{2}$ & $1.206 \pm 0.321$\,s \\
 & CMA-ES & $1.505\times 10^{3} \pm 1.717\times 10^{3}$ & $1.126 \pm 0.490$\,s \\
 & FCPO & $\textbf{$9.625\times 10^{2} \pm 1.275\times 10^{3}$}$ & $0.602 \pm 0.011$\,s \\
 & FCPO\_NoZoom & $1.757\times 10^{3} \pm 9.536\times 10^{2}$ & $0.550 \pm 0.233$\,s \\
 & FCPO\_NoEigen & $1.329\times 10^{3} \pm 1.451\times 10^{3}$ & $0.585 \pm 0.011$\,s \\
 & FCPO\_NoLPSR & $1.233\times 10^{3} \pm 1.267\times 10^{3}$ & $0.976 \pm 0.243$\,s \\
\bottomrule
\end{tabular}
\end{table}

\begin{table}[!t]
\centering
\scriptsize
\caption{Aggregate performance across the 10 benchmark cases (5 functions $\times$ 2 dimensions). Avg. rank denotes the mean rank (lower is better), computed from the per-case median final objective value. Mean time denotes the average of the per-case mean runtimes, in seconds.}
\label{tab:agg_rank_time}
\begin{tabular}{lcc}
\toprule
Algorithm & Avg. rank & Mean time (s) \\
\midrule
CMA-ES & 2.20 & 0.425 \\
FCPO\_NoLPSR & 3.15 & 0.288 \\
FCPO & 4.90 & 0.183 \\
FCPO\_NoEigen & 4.90 & 0.187 \\
L-SHADE & 5.10 & 0.480 \\
FCPO\_NoZoom & 5.25 & 0.180 \\
CLPSO & 5.90 & 0.346 \\
SHADE & 5.90 & 0.479 \\
CSO & 8.10 & 0.412 \\
PSO & 9.60 & 0.274 \\
\bottomrule
\end{tabular}
\end{table}

\end{document}